# Second-Order Consistencies

**Christophe Lecoutre**                 LECOUTRE@CRIL.FR
**Stéphane Cardon**                   CARDON@CRIL.FR
*CRIL-CNRS UMR 8188*
*Université Lille-Nord de France, Artois*
*rue de l'université*
*SP 16, F-62307 Lens, France*

**Julien Vion**               JULIEN.VION@UNIV-VALENCIENNES.FR
*LAMIH-CNRS FRE 3304*
*Université Lille-Nord de France, UVHC*
*F-59313 Valenciennes Cedex 9, France*

## Abstract

In this paper, we propose a comprehensive study of second-order consistencies (i.e., consistencies identifying inconsistent pairs of values) for constraint satisfaction. We build a full picture of the relationships existing between four basic second-order consistencies, namely path consistency (PC), 3-consistency (3C), dual consistency (DC) and 2-singleton arc consistency (2SAC), as well as their conservative and strong variants. Interestingly, dual consistency is an original property that can be established by using the outcome of the enforcement of generalized arc consistency (GAC), which makes it rather easy to obtain since constraint solvers typically maintain GAC during search. On binary constraint networks, DC is equivalent to PC, but its restriction to existing constraints, called conservative dual consistency (CDC), is strictly stronger than traditional conservative consistencies derived from path consistency, namely partial path consistency (PPC) and conservative path consistency (CPC). After introducing a general algorithm to enforce strong (C)DC, we present the results of an experimentation over a wide range of benchmarks that demonstrate the interest of (conservative) dual consistency. In particular, we show that enforcing (C)DC before search clearly improves the performance of MAC (the algorithm that maintains GAC during search) on several binary and non-binary structured problems.

## 1. Introduction

Many decision problems are combinatorial by nature, and can be modelled using finite domain variables connected with constraints. Such models are formally represented as *constraint networks* (CNs) and finding a solution to a model is an instance of the NP-complete *constraint satisfaction problem* (CSP). The CSP is usually solved through systematic *backtrack search*, a fundamental technique in artificial intelligence. There have been considerable efforts during the last three decades to improve the practical efficiency of backtrack search.

*Consistencies* are properties of constraint networks that can be exploited (enforced), before or during search, to filter the search space of problem instances by inference. Currently, the most successful consistencies are *domain filtering* consistencies (Debruyne & Bessiere, 2001; Bessiere, Stergiou, & Walsh, 2008). Common consistencies for binary CNs are *arc consistency* (AC, Mackworth, 1977) and *singleton arc consistency* (SAC, Bessiere & De-bruyne, 2005). Example consistencies for non-binary CNs are *generalized arc consistency* (GAC, Mohr & Masini, 1988) and *pairwise inverse consistency* (PWIC, Stergiou & Walsh, 2006). Consistencies typically allow the identification of nogoods. A *nogood* is an instantiation of some variables that cannot lead to any solution. Identifying relevant nogoods as soon as possible when exploring the search space of an instance is recognized as an essential component of a backtracking search algorithm. A domain-filtering consistency is also called a first-order consistency, meaning that it only detects *inconsistent* values (1-sized nogoods): these values can be safely removed from the domains of the variables.

This paper is concerned with *second-order* consistencies that locally identify inconsistent pairs of values. The most studied second-order consistency is *path consistency* (PC, Montanari, 1974; Mackworth, 1977). For now, path consistency, and more generally higher order consistencies, are rather neglected by designers and developers of general constraint solvers. This is somewhat surprising since, for many tractable classes, strong path consistency (path consistency combined with arc consistency) is a sufficient condition to determine satisfiability (e.g., see Dechter, 1992; van Beek, 1992; Cooper, Cohen, & Jeavons, 1994; Zhang & Yap, 2006). Neglecting higher order consistencies may partly be due to the somewhat limited scope of these classes: exciting progress in this area has only been very recent (e.g., see Green & Cohen, 2008). However, path consistency has an important role in temporal reasoning. Indeed, for some classes of interval algebra, path consistency – adapted to temporal constraint networks (Allen, 1983) – is sufficient to decide satisfiability. Another possible reason for the low practical interest for path consistency, in the discrete constraint satisfaction field, is that path consistency enforcement modifies constraint relations, and more importantly, modifies the structure of the constraint graph. When a pair of values $(a, b)$ for the variables $(x, y)$ is found to be path-inconsistent, this information is recorded within the constraint network; if there is no constraint binding $x$ with $y$, a new one is inserted, thus changing the constraint graph. For example, the instance scen-11 of the radio link frequency assignment problem (Cabon, de Givry, Lobjois, Schiex, & Warners, 1999) involves 680 variables and 4,103 constraints. Enforcing a second-order consistency on this network could at worst create $\binom{680}{2} - 4{,}103 = 226{,}757$ new constraints, which would be really counter-productive both in time and in space. The main apparent drawback of path consistency can be avoided by adopting a *conservative* approach, in which the search for inconsistent pairs of values is restricted to existing constraints. This is called *conservative path consistency* (CPC, Debruyne, 1999) when restricted to paths of length 2 in the constraint graph, and *partial path consistency* (PPC, Bliek & Sam-Haroud, 1999) when restricted to paths of arbitrary length in the constraint graph; CPC and PPC are equivalent when the constraint graph is triangulated.

In this paper, we study path consistency as well as three other basic second-order consistencies that are 3-consistency (3C, Freuder, 1978), dual consistency (DC, Lecoutre, Cardon, & Vion, 2007a) and 2-singleton arc consistency (2SAC, Bessiere, Coletta, & Petit, 2005). On binary constraint networks, DC is equivalent to PC – McGregor (1979) proposed an DC-like algorithm to establish (strong) path consistency – but when considering weaker *conservative* variants, we show that *conservative dual consistency* (CDC) is strictly stronger than PPC and CPC: CDC can filter out more inconsistent pairs of values (from existing constraints) than PPC or CPC. We build a full picture of the qualitative relationships existing between all those second-order consistencies (including the stronger 2SAC property, the conserva-tive restrictions and strong variants of all studied consistencies) for both binary CNs and non-binary CNs.

Interestingly enough, (conservative) dual consistency benefits from some nice features: (1) as (C)DC is built on top of GAC, implementing a filtering algorithm to enforce it is rather easy, (2) for the same reason, all optimizations achieved on GAC algorithms these last years come for free, (3) we have the guarantee that GAC enforced on a CN that verifies the property (C)DC leaves the property unchanged. This is why our theoretical study is followed by the presentation of a general algorithm to enforce strong (C)DC and an experimental study to show the practical interest of using (C)DC (during a preprocessing step) when solving binary and non-binary problem instances with a search algorithm such as MAC (Sabin & Freuder, 1994).

The paper is organized as follows. Section 2 introduces technical background about constraint networks, nogoods and consistencies. In Section 3, we introduce (basic, conservative and strong) second-order consistencies, with a focus on path consistency and a possible misunderstanding about it. A qualitative study about second-order consistencies is conducted in Section 4. An algorithm to enforce (C)DC is proposed in Section 5, and experimental results are presented in Section 6. Finally, we conclude.

## 2. Technical Background

This section provides technical background about constraint networks and consistencies.

### 2.1 Constraint Networks

A (finite) constraint network (CN) $P$ is composed of a finite set of $n$ variables, denoted by vars($P$), and a finite set of $e$ constraints, denoted by cons($P$). Each *variable* $x$ has an associated *domain*, denoted by dom($x$), that contains the finite set of values that can be assigned to $x$. Each *constraint* $c$ involves an ordered set of variables, called the *scope* of $c$ and denoted by scp($c$). It is defined by a *relation*, denoted by rel($c$), which contains the set of tuples allowed for the variables involved in $c$. The *arity* of a constraint $c$ is the size of scp($c$). The maximum domain size and the maximum arity for a given CN will be denoted by $d$ and $r$, respectively. A *binary* constraint involves exactly 2 variables, and a *non-binary* constraint strictly more than 2 variables. A binary CN only contains binary constraints whereas a non-binary CN contains at least one non-binary constraint.

The *initial* domain of a variable $x$ is denoted by $\mathrm{dom}^{\mathrm{init}}(x)$ whereas the *current* domain of $x$ in the CN $P$ is denoted by $\mathrm{dom}^P(x)$ or more simply dom($x$) when the context is unambiguous. The initial relation of a constraint $c$ is denoted by $\mathrm{rel}^{\mathrm{init}}(c)$ whereas the current relation is denoted by $\mathrm{rel}^P(c)$ or more simply rel($c$). A constraint $c$ is *universal* iff $\mathrm{rel}^{\mathrm{init}}(c) = \Pi_{x \in \mathrm{scp}(c)} \mathrm{dom}^{\mathrm{init}}(x)$; a universal constraint imposes no restriction. We consider that for any variable $x$, we always have dom($x$) $\subseteq \mathrm{dom}^{\mathrm{init}}(x)$, and for any constraint $c$, we always have rel($c$) $\subseteq \mathrm{rel}^{\mathrm{init}}(c)$. To simplify, a pair $(x, a)$ with $x \in \mathrm{vars}(P)$ and $a \in \mathrm{dom}(x)$ is called a (current) *value of $P$*. Without any loss of generality, we only consider CNs that involve neither unary constraints (i.e., constraints involving a unique variable) nor constraints of similar scope – CNs are *normalized* (Apt, 2003; Bessiere, 2006). The set of normalized CNs with neither unary constraints nor universal constraints[1] is denoted by $\mathscr{P}$.

An *instantiation* $I$ of a set $X = \{x_1, \ldots, x_k\}$ of variables is a set $\{(x_1, a_1), \ldots, (x_k, a_k)\}$ such that $\forall i \in 1..k, a_i \in \text{dom}^{\text{init}}(x_i)$ ; the set $X$ of variables occurring in $I$ is denoted by $\text{vars}(I)$ and each value $a_i$ is denoted by $I[x_i]$. An instantiation $I$ on a CN $P$ is an instantiation of a set $X \subseteq \text{vars}(P)$; it is *complete* iff $\text{vars}(I) = \text{vars}(P)$, *partial* otherwise. $I$ is *valid* on $P$ iff $\forall (x, a) \in I, a \in \text{dom}^P(x)$. An instantiation $I$ *covers* a constraint $c$ iff $\text{scp}(c) \subseteq \text{vars}(I)$, and *satisfies* a constraint $c$ with $\text{scp}(c) = \{x_1, \ldots, x_r\}$ iff (1) $I$ covers $c$ and (2) the tuple $(a_1, \ldots, a_r)$ is allowed by $c$, i.e., $(a_1, \ldots, a_r) \in \text{rel}(c)$, where $\forall i \in 1..r, a_i = I[x_i]$. A *support* (resp., a *conflict*) on a constraint $c$ is a valid instantiation of $\text{scp}(c)$ that satisfies (resp., does not satisfy) $c$. An instantiation $I$ on a CN $P$ is *locally consistent* iff (1) $I$ is valid on $P$ and (2) every constraint of $P$ covered by $I$ is satisfied by $I$. It is *locally inconsistent* otherwise. A *solution* of $P$ is a complete instantiation on $P$ that is locally consistent. An instantiation $I$ on a CN $P$ is *globally inconsistent*, or a *nogood*, iff it cannot be extended to a solution of $P$. It is *globally consistent* otherwise. We refer here to standard nogoods (e.g., see Dechter, 2003); they differ from nogoods coming with justifications (Schiex & Verfaillie, 1994) and from generalized ones (Katsirelos & Bacchus, 2003). Two CNs $P$ and $P'$ defined on the same variables are *equivalent* iff they have the same solutions.

A CN is said to be *satisfiable* iff it admits at least one solution. The Constraint Satisfaction Problem (CSP) is the NP-complete task of determining whether a given CN is satisfiable or not. Thus, a CSP instance is defined by a CN which is solved either by finding a solution or by proving unsatisfiability. In many cases, a CSP instance can be solved by using a combination of search and inferential simplification (Dechter, 2003; Lecoutre, 2009). To solve a CSP instance, a depth-first search algorithm with backtracking can be applied, where at each step of the search, a variable assignment is performed followed by a filtering process called *constraint propagation*. Constraint propagation algorithms enforce some consistency property, they identify and record explicit nogoods in CNs. When identified nogoods are of size 1, they correspond to inconsistent values.

It is usual to refer to some properties of the (hyper)graphs that can be associated with any CN. On the one hand, the *constraint (hyper)graph*, also called *macro-structure*, associated with a (normalized) CN $P$ consists of $n$ vertices corresponding to the variables of $P$ and also $e$ (hyper)edges corresponding to the constraints of $P$: an (hyper)edge connects vertices corresponding to the variables in the scope of the constraint it represents. On the other hand, the *compatibility (hyper)graph*, also called *micro-structure* (Jégou, 1993), associated with a normalized CN $P$ contains one vertex per value of $P$ and one (hyper)edge per constraint support. It corresponds to a $n$-partite hypergraph with one part for each variable. Sometimes, incompatibility (hyper)graphs are used by authors where (hyper)edges correspond to conflicts instead of supports. In this paper, (hyper)edges for supports (resp., conflicts) will be drawn using solid (resp., dashed) lines.

It is sometimes helpful to use a homogeneous representation of a CN, wherein domains and also constraints are replaced by nogoods. The *nogood representation* of a CN is a set of nogoods, one for every value removed from the initial domain of a variable and one for every tuple forbidden by a constraint. More precisely, the *nogood representation* $\widetilde{x}$ of a variable $x$ is the set of instantiations $\{\{(x, a)\} \mid a \in \mathrm{dom}^{\mathrm{init}}(x) \setminus \mathrm{dom}(x)\}$. The *nogood representation* $\widetilde{c}$ of a constraint $c$, with $\mathrm{scp}(c) = \{x_1, \ldots, x_r\}$, is the set of instantiations $\left\{ \{(x_1, a_1), \ldots, (x_r, a_r)\} \mid (a_1, \ldots, a_r) \in \left( \prod_{x \in \mathrm{scp}(c)} \mathrm{dom}^{\mathrm{init}}(x) \right) \setminus \mathrm{rel}(c) \right\}$. The *nogood representation* $\widetilde{P}$ of a CN $P$ is the set of instantiations $\left( \bigcup_{x \in \mathrm{vars}(P)} \widetilde{x} \right) \cup \left( \bigcup_{c \in \mathrm{cons}(P)} \widetilde{c} \right)$.

---

1. For our theoretical study, universal constraints can be safely ignored. In practice, any universal constraint $c$ in a CN $P$ may artificially be considered as non-universal: choose a variable $x \in \text{scp}(c)$ and consider a dummy value $\nu$ such that $\nu \in \text{dom}^{\text{init}}(x) \setminus \text{dom}^P(x)$ and $\text{rel}^{\text{init}}(c)$ forbids one tuple involving $(x, \nu)$.

Instantiations in $\widetilde{P}$ are *explicit nogoods* of $P$ (recorded through domains and constraints). Notice that when a nogood is a superset of another one, it is *subsumed*. Intuitively, a nogood that is subsumed is not relevant as it is less general than at least another one, and two CNs are *nogood-equivalent* – a related definition is Definition 3.11 in the work of Bessiere (2006) – when they have the same canonical form, i.e., represent exactly the same set of "unsubsumed" nogoods. To relate CNs, we introduce a general partial order.[2] Let $P$ and $P'$ be two CNs defined on the same variables (i.e., such that $\mathrm{vars}(P) = \mathrm{vars}(P')$), $P' \preceq P$ iff $\widetilde{P'} \supseteq \widetilde{P}$ and $P' \prec P$ iff $\widetilde{P'} \supsetneq \widetilde{P}$. $(\mathscr{P}, \preceq)$ is a partially ordered set (poset) because $\preceq$ is reflexive, antisymmetric (remember that no CN in $\mathscr{P}$ can involve universal constraints) and transitive. As CNs are normalized and no unary or universal constraint is present, there is therefore only one manner to discard (or remove) an instantiation from a given CN, or equivalently to "record" a new explicit nogood in a CN. Given a CN $P$ in $\mathscr{P}$, and an instantiation $I$ on $P$, $P \setminus I$ denotes the CN $P'$ in $\mathscr{P}$ such that $\mathrm{vars}(P') = \mathrm{vars}(P)$, and $\widetilde{P'} = \widetilde{P} \cup \{I\}$. $P \setminus I$ is an operation that retracts $I$ from $P$ and builds a new CN, not necessarily with the same set of constraints. Let us show how $P'$ is built. If $I \in \widetilde{P}$, of course we have $P' = P \setminus I = P$: this means that the instantiation $I$ was already an explicit nogood of $P$. The interesting case is when $I \notin \widetilde{P}$. If $I$ corresponds to a value $a$ for a variable $x$, i.e., $I = \{(x, a)\}$, it suffices to remove $a$ from $\mathrm{dom}(x)$. If $I$ corresponds to a tuple allowed by a constraint $c$ of $P$, it suffices to remove this tuple from $\mathrm{rel}(c)$. Otherwise, we must introduce a new constraint whose associated relation contains all possible tuples (built from initial domains) except the one that corresponds to the instantiation $I$. Note that removing a tuple from a relation $\mathrm{rel}(c)$ can be a problem in practice if the constraint $c$ is defined in intension (i.e., by a predicate). However, for binary nogoods (our concern), this is not a real problem because, except when variables have very large domains, it is always possible to translate (efficiently) an intensional constraint in extension.

## 2.2 Consistencies

A *consistency* is a general property of a CN. When a consistency $\phi$ holds on a CN $P$, we say that $P$ is $\phi$-consistent. If $\phi$ and $\psi$ are two consistencies, a CN $P$ is said to be $\phi + \psi$-*consistent* iff $P$ is both $\phi$-consistent and $\psi$-consistent. A consistency $\phi$ is *nogood-identifying* iff the reason why a CN $P$ is not $\phi$-consistent is that some instantiations, which are not in $\widetilde{P}$, are identified as globally inconsistent by $\phi$. Such instantiations correspond to (new identified) nogoods and are said to be $\phi$-*inconsistent* (on $P$). A *$k$th-order consistency* is a nogood-identifying consistency that allows the identification of nogoods of size $k$, where $k \geq 1$ is an integer. $k$th-order consistency should not be confused with $k$-consistency (Freuder, 1978, 1982): $k$-consistency holds iff every locally consistent instantiation of a set of $k-1$ variables can be extended to a locally consistent instantiation involving any additional variable. In our terminology, this is a $(k-1)$th-order consistency.

---

2. This partial order is general enough for our purpose, but note that more sophisticated partial orders (or preorders) exist (e.g., by taking account of subsumed nogoods).

A *domain-filtering consistency* is a first-order consistency. A *conservative consistency* $\phi$ is a nogood-identifying consistency such that, for every given CN $P$, every $\phi$-inconsistent instantiation on $P$ corresponds to a tuple currently allowed by an explicit constraint of $P$. To compare the pruning capability of consistencies, we introduce a preorder (see Debruyne & Bessiere, 2001). A consistency $\phi$ is *stronger* than (or equal to) $\psi$ iff whenever $\phi$ holds on a CN $P$, $\psi$ also holds on $P$. $\phi$ is *strictly stronger* than $\psi$ iff $\phi$ is stronger than $\psi$ and there exists at least one CN $P$ such that $\psi$ holds on $P$ but not $\phi$. When some consistencies cannot be ordered (none is stronger that another), we say that they are *incomparable*.

We now briefly introduce a formal characterization of constraint propagation, based on the concept of stability (following Lecoutre, 2009). This formalism is also related to previous works about local consistencies and rules iteration (e.g., see Montanari & Rossi, 1991; Apt, 1999, 2003; Bessiere, 2006). It is usually possible to enforce $\phi$ on a CN $P$ by computing the greatest $\phi$-consistent CN smaller than or equal to $P$, while preserving the set of solutions. A consistency is *well-behaved* when for any CN $P \in \mathscr{P}$, the set $\{P' \in \mathscr{P} \mid P'$ is $\phi$-consistent and $P' \preceq P\}$ admits a greatest element, denoted by $\phi(P)$, that is equivalent to $P$ and called the *$\phi$-closure* of $P$. Enforcing $\phi$ on a CN $P$ means computing $\phi(P)$, and an algorithm that enforces $\phi$ is called a *$\phi$-algorithm*. The property of stability is useful for proving that a nogood-identifying consistency is well-behaved.

A nogood-identifying consistency $\phi$ is *stable* iff for every CN $P \in \mathscr{P}$, every CN $P' \in \mathscr{P}$ such that $P' \preceq P$ and every $\phi$-inconsistent instantiation $I$ on $P$, we have either $I \in \widetilde{P'}$ or $I$ is $\phi$-inconsistent on $P'$; the second condition for stability given by Lecoutre (2009) holds necessarily because of the choice of the poset in this paper. The fact that either $I \in \widetilde{P'}$ or $I$ is $\phi$-inconsistent on $P'$ guarantees that no $\phi$-inconsistent instantiation on a CN can be missed when the CN is made tighter: either it is discarded (has become an explicit nogood of $P'$) or it remains $\phi$-inconsistent.

**Theorem 1.** *(Lecoutre, 2009) Any stable nogood-identifying consistency is well-behaved.*

The stability of a nogood-identifying consistency $\phi$ provides a general procedure for computing the $\phi$-closure of any CN: iteratively discard (in any order) $\phi$-inconsistent instantiations until a fixed point is reached. Provided that the procedure is sound (each removal corresponds to a $\phi$-inconsistent instantiation) and complete (each $\phi$-inconsistent instantiation is removed), the procedure is guaranteed to compute $\phi$-closures. More generally, when different reduction rules are used, each must be shown to be correct, monotonic and inflationary. We can then benefit from a generic iteration algorithm (Apt, 2003, Lemmas 7.5, 7.8 and Theorem 7.11). Stability under union can also be proved for a domain-filtering consistency, thus guaranteeing a fixed point (Bessiere, 2006). An interesting result follows:

**Proposition 1.** *Let $\phi$ and $\psi$ be two well-behaved (nogood-identifying) consistencies. $\phi$ is stronger than $\psi$ iff for every CN $P \in \mathscr{P}$, we have $\phi(P) \preceq \psi(P)$.*

We conclude this section with some well-known domain-filtering consistencies. First, let us introduce generalized arc consistency (GAC). A *support* (resp., a *conflict*) *for* a value $(x, a)$ of $P$ on a constraint $c$ involving $x$ is a support (resp., a conflict) $I$ on $c$ such that $I[x] = a$. A value $(x, a)$ of $P$ is GAC-consistent iff there exists a support for $(x, a)$ on every constraint of $P$ involving $x$. $P$ is GAC-consistent iff every value of $P$ is GAC-consistent. We also say that a constraint $c$ is GAC-consistent iff for every variable $x$ in $\mathrm{scp}(c)$ and every value $a$ in $\mathrm{dom}(x)$, $(x, a)$ is GAC-consistent. For binary CNs, GAC is referred to as AC (Arc Consistency). Second, we introduce some "singleton" consistencies (Debruyne & Bessiere, 1997b; Prosser, Stergiou, & Walsh, 2000). When the domain of a variable of $P$ is empty, $P$ is clearly unsatisfiable, which is denoted by $P = \bot$. The CN $P|_{x=a}$ is obtained from $P$ by removing every value $b \neq a$ from $\mathrm{dom}(x)$. A value $(x, a)$ of $P$ is SAC-consistent (SAC stands for Singleton Arc Consistent[3]) iff $GAC(P|_{x=a}) \neq \bot$ (this is called a *singleton check*). A value $(x, a)$ of $P$ is BiSAC-consistent iff $GAC(P^{ia}|_{x=a}) \neq \bot$, where $P^{ia}$ is the CN obtained after removing every value $(y, b)$ of $P$ such that $y \neq x$ and $(x, a) \notin GAC(P|_{y=b})$ (Bessiere & Debruyne, 2008). $P$ is SAC-consistent (respectively, BiSAC-consistent) iff every value of $P$ is SAC-consistent (respectively, BiSAC-consistent). BiSAC is strictly stronger than SAC, which is itself strictly stronger than GAC; BiSAC is also strictly weaker than strong path consistency (Bessiere & Debruyne, 2008). GAC, SAC and BiSAC are well-behaved; for example, $SAC(P)$ denotes the SAC-closure of the CN $P$.

## 3. Second-Order Consistencies

In this section, we introduce second-order consistencies. First, we start with the most famous one: path consistency. Then, we clarify some aspects of path consistency that are sometimes misrepresented in the literature, and introduce its known restricted forms. Finally, we introduce 3-consistency, dual consistency, and 2-singleton arc consistency as well as their conservative and strong variants.

### 3.1 Path Consistency

Among the consistencies that allow us to identify inconsistent pairs of values, *path consistency* plays a central role. Introduced by Montanari (1974), its definition has sometimes been misinterpreted. The problem arises around the definition of a "path", which must be understood as any sequence of variables, and not as a sequence of variables that corresponds to a path in the constraint graph. This ambiguity probably comes from Montanari's original paper, in which reasoning from path consistency is achieved with respect to complete (or completion of) constraint graphs, although a footnote in the original paper indicates:

> "A path in a network is any sequence of vertices. A vertex can occur more than once in a path even in consecutive positions."

A precise definition of *path* is thus required. The definition of path below is used by Montanari (1974), Mackworth (1977) and Debruyne (1998), while the definition of graph-path is used, for example, by Tsang (1993) or Bessiere (2006). A path is an arbitrary sequence of variables, and a *graph-path* is defined to be a sequence of variables such that a binary constraint exists between any two variables adjacent in the sequence. For a binary CN $P$, a graph-path is thus a path in the constraint graph of $P$. For a non-binary CN $P$, if all non-binary constraints are discarded (ignored) then a path in the resulting constraint graph is a graph-path. It is important to note that any given variable may occur several times in a path (and so, in a graph-path). Figure 1 gives an illustration.

---

3. To limit the number of acronyms, we use SAC for both binary and non-binary CNs.

**Definition 1** (Path). *Let $P$ be a CN.*

**A *path* of** $P$ *is a sequence $\langle x_1, \ldots, x_k \rangle$ of variables of $P$ such that $x_1 \neq x_k$ and $k \geq 2$; the path is from variable $x_1$ to variable $x_k$, and $k - 1$ is the length of the path.*

**A *graph-path* of** $P$ *is a path $\langle x_1, \ldots, x_k \rangle$ of $P$ such that $\forall i \in 1..k-1, \exists c \in \mathrm{cons}(P) \mid \mathrm{scp}(c) = \{x_i, x_{i+1}\}$.*

**A *closed (graph-)path* of** $P$ *is a (graph-)path $\langle x_1, \ldots, x_k \rangle$ of $P$ such that $\exists c \in \mathrm{cons}(P) \mid \mathrm{scp}(c) = \{x_1, x_k\}$.*

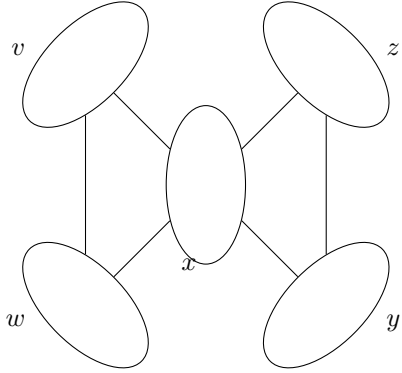

Figure 1: The constraint graph of a binary CN $P$. $\langle v, z, x \rangle$ and $\langle v, y, w, y \rangle$ are paths of $P$. $\langle v, z, y, w \rangle$ is a closed path of $P$. $\langle v, x, y \rangle$ is a graph-path of $P$. $\langle v, x, w \rangle$ and $\langle z, x, v, x, y \rangle$ are two closed graph-paths of $P$.

The central concept of *consistent paths* is defined as follows:

**Definition 2** (Consistent Path). *Let $P$ be a CN.*

- *An instantiation $\{(x_1, a_1), (x_k, a_k)\}$ on $P$ is consistent on a path $\langle x_1, \ldots, x_k \rangle$ of $P$ iff there exists a tuple $\tau \in \Pi_{i=1}^{k} \mathrm{dom}(x_i)$ such that $\tau[x_1] = a_1$, $\tau[x_k] = a_k$ and $\forall i \in 1..k-1, \{(x_i, \tau[x_i]), (x_{i+1}, \tau[x_{i+1}])\}$ is a locally consistent instantiation[4] on $P$. The tuple $\tau$ is said to be a support for $\{(x_1, a_1), (x_k, a_k)\}$ on $\langle x_1, \ldots, x_k \rangle$ (in $P$).*

- *A path $\langle x_1, \ldots, x_k \rangle$ of $P$ is consistent iff every locally consistent instantiation of $\{x_1, x_k\}$ on $P$ is consistent on $\langle x_1, \ldots, x_k \rangle$.*

In the example in Figure 2, $\langle v, z, x \rangle$ is a consistent path of $P$ since for the locally consistent instantiation $\{(v, a), (x, b)\}$ we can find $b$ in $\mathrm{dom}(z)$ such that $\{(v, a), (z, b)\}$ is locally consistent (this is trivial since there is an implicit universal binary constraint between $v$ and $z$) and $\{(x, b), (z, b)\}$ is locally consistent. Similarly, the second locally consistent instantiation $\{(v, b), (x, a)\}$ can be extended to $z$. The closed graph-path $\langle v, x, w \rangle$ is not consistent; the locally consistent instantiation $\{(v, b), (w, a)\}$ cannot be extended to $x$. One might be surprised that $\langle z, x, v, w, x, y \rangle$ is consistent. It is important to note that we are free to select different values for $x$ along the path. For example, for the locally consistent instantiation $\{(z, b), (y, a)\}$, we can find the support $\tau = (b, b, a, b, a, a)$ on $\langle z, x, v, w, x, y \rangle$. This tuple belongs to $\mathrm{dom}(z) \times \mathrm{dom}(x) \times \mathrm{dom}(v) \times \mathrm{dom}(w) \times \mathrm{dom}(x) \times \mathrm{dom}(y)$, satisfies $\tau[z] = b$, $\tau[y] = a$ and all encountered binary constraints along the path. Along this path we have first $(x, b)$ and subsequently $(x, a)$.

---

4. If $x_i = x_{i+1}$, then necessarily $\tau[x_i] = \tau[x_{i+1}]$ because an instantiation cannot contain two distinct pairs involving the same variable.

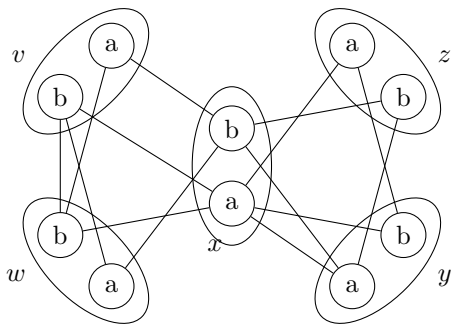

Figure 2: The compatibility graph of a binary CN $P$ (whose constraint graph is given by Figure 1). $\langle v, z, x \rangle$ is a consistent path of $P$. The closed graph-path $\langle v, x, w \rangle$ is not consistent contrary to $\langle z, x, v, w, x, y \rangle$.

We can now introduce the historical definition of *path consistency* (PC, Montanari, 1974; Mackworth, 1977).

**Definition 3** (Path Consistency). *A CN $P$ is* path-consistent, *denoted* PC-consistent, *iff every path of $P$ is consistent.*

This definition is valid for non-binary CNs. Simply, non-binary constraints are ignored, as Dechter (2003) or Bessiere (2006) do, since in Definition 2, only pairs of variables are considered. Montanari has shown that it is sufficient to consider paths of length two (i.e., sequences of three variables) only. Note that it is not necessary for the constraint graph to be complete (but, when path consistency is enforced, the resulting CN may become complete).

**Theorem 2.** *(Montanari, 1974) A CN $P$ is path-consistent iff every 2-length path of $P$ (i.e., every sequence of three variables) is consistent.*

This leads to the following classical definition:

**Definition 4** (Path Consistency). *Let $P$ be a CN.*

- *An instantiation[5] $\{(x, a), (y, b)\}$ on $P$ is* path-consistent, *denoted* PC-consistent, *iff it is 2-length path-consistent, that is to say, iff there exists a value $c$ in the domain of every third variable $z$ of $P$ such that $\{(x, a), (z, c)\}$ and $\{(y, b), (z, c)\}$ are both locally*

---

5. In this paper, when we refer to an instantiation of the form $\{(x, a), (y, b)\}$, we assume that $x \neq y$.

*consistent; if $\{(x, a), (y, b)\}$ is not path-consistent, it is said to be path-inconsistent or PC-inconsistent.*

- *$P$ is path-consistent iff every locally consistent instantiation $\{(x, a), (y, b)\}$ on $P$ is path-consistent.*

### 3.2 Deep in Path Consistency

Now, we show that path consistency may be easily misinterpreted, and we introduce consistency forms related to PC. A first natural question is: can we restrict our attention to graph-paths (see Definition 1)? The answer is given by the following observation.

**Observation 1.** *For some CNs, the following properties are not equivalent:*

(a) *every path is consistent*

(b) *every graph-path is consistent*

*Proof.* Consider, for example, the CN depicted in Figure 3. This CN is not path-consistent since the locally consistent instantiation $\{(x, b), (z, a)\}$ is not consistent on the path $\langle x, y, z \rangle$. If we now limit our attention to graph-paths, these consist only of the variables $x$ and $y$ (for example, $\langle x, y \rangle$, $\langle x, y, x, y \rangle$, ...) and there is no local inconsistency. □

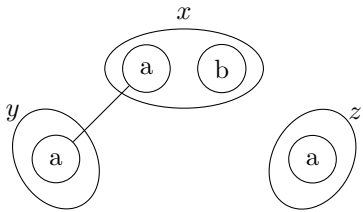

Figure 3: A CN $P$ with three variables and only one constraint (between $x$ and $y$). $P$ is not path-consistent. However, any graph-path that can be built is consistent.

However, for a binary CN that has a connected constraint graph (i.e., a constraint graph composed of a single connected component), the restriction to graph-paths is valid.

**Proposition 2.** *Let $P$ be a binary CN such that $P \neq \perp$ and the constraint graph of $P$ is connected. $P$ is path-consistent iff every graph-path of $P$ is consistent.*

*Proof.* For one direction ($\Rightarrow$), this is immediate. If $P$ is path-consistent, then by definition every path of $P$ is consistent, including graph-paths.

For the other direction ($\Leftarrow$), we show that if every graph-path of $P$ is consistent, then every 2-length path of $P$ is consistent (and thus $P$ is path-consistent using Theorem 2). In practical terms, we consider a locally consistent instantiation $\{(x, a), (y, b)\}$ and show that for each third variable $z$ of $P$, the following property $Pr(z)$ holds: $\exists c \in \text{dom}(z)$ such that $\{(x, a), (z, c)\}$ and $\{(y, b), (z, c)\}$ are both locally consistent instantiations. For each variable $z$ three cases must be considered, depending of the existence of the constraints $c_{xz}$, between $x$ and $z$ (i.e., such that $\text{scp}(c_{xz}) = \{x, z\}$), and $c_{yz}$, between $y$ and $z$.

(a) Both constraints exist: thus there exists a graph-path $\langle x, z, y \rangle$ and as this path is consistent by hypothesis, the property $Pr(z)$ holds.

(b) Neither constraint exist: $P \neq \bot$ implies $\text{dom}(z) \neq \emptyset$, thus $Pr(z)$ holds because $c_{xz}$ and $c_{yz}$ are implicit and universal.

(c) Only the constraint $c_{xz}$ exists (similarly, only the constraint $c_{yz}$ exists): as the constraint graph is connected, there exists at least one graph-path from $z$ to $y$, and consequently a graph-path from $x$ to $y$ of the form $\langle x, z, \ldots, y \rangle$. This means that there is a value in $\text{dom}(z)$ which is compatible with $(x, a)$, by using the hypothesis (every graph-path is consistent). This value is also compatible with $(y, b)$ because there is an implicit universal constraint between $z$ and $y$. Hence, $Pr(z)$ holds. □

Unsurprisingly (because of Observation 1), a binary CN $P$ such that every 2-length graph-path of $P$ is consistent, is not necessarily path-consistent. Of course, in the special case where the constraint graph is complete, the CN is path-consistent because every path of $P$ is also a graph-path of $P$.

**Observation 2.** *For some CNs, the following properties are not equivalent:*

*(a) every graph-path is consistent*

*(b) every 2-length graph-path is consistent*

*Proof.* See Figure 4. □

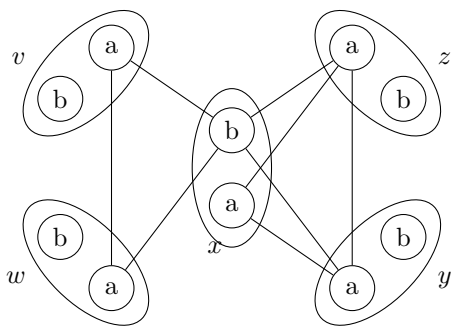

Figure 4: Every 2-length graph-path of this CN (whose constraint graph is given by Figure 1) is consistent. The graph-path $\langle x, w, v, x, z \rangle$ is not consistent for $\{(x, a), (z, a)\}$.

We have the following proposition for 2-length graph-paths.

**Proposition 3.** *Let $P$ be a binary CN such that $P \neq \bot$ and $P$ is arc-consistent. $P$ is path-consistent iff every 2-length graph-path of $P$ is consistent.*

*Proof.* The proof is similar to the proof of Proposition 2 by considering 2-length graph-paths instead of graph-paths. Only case (c) in the demonstration differs.

(c) Only the constraint $c_{xz}$ exists (similarly, only the constraint $c_{yz}$ exists): as $P$ is arc-consistent, there exists a value in $\text{dom}(z)$ that is compatible with $(x, a)$. Because there is an implicit universal constraint between $z$ and $y$, this value is also compatible with $(y, b)$. Hence, $Pr(z)$ holds. □

Theorem 2 and Propositions 2 and 3 suggest that the historical definition of path consistency is appropriate since it corresponds to the strongest form of detection of local inconsistencies using the concept of path (even if considering graph-paths may seem more natural than considering paths). Unfortunately, path consistency has sometimes been misinterpreted. For example, Definition 3-11 in a work of Tsang (1993) links PC to graph-paths, and Proposition 3.39 in a work of Bessiere (2006) links PC to 2-length graph-paths, but Observations 1 and 2 (with Figure 4) show that this is not equivalent to Montanari's definition. However, it is true that to check path consistency in practice, we only need to consider 2-length graph-paths, provided that binary constraints are arc-consistent.

Two different relation-filtering consistencies related to path consistency have been defined in terms of closed graph-paths. The first is *partial path consistency* (partial PC or PPC, Bliek & Sam-Haroud, 1999) and the second is *conservative path consistency* (conservative PC or CPC, Debruyne, 1999).

**Definition 5** (Partial Path Consistency). *A CN $P$ is partially path-consistent, denoted* PPC-consistent, *iff every closed graph-path of $P$ is consistent.*

**Definition 6** (Conservative Path Consistency). *A CN $P$ is conservative path-consistent, denoted* CPC-consistent, *iff every closed 2-length graph-path of $P$ is consistent.*

For binary constraints, PPC and CPC are equivalent when the constraint graph is triangulated: PPC was initially introduced to build a filtering algorithm that operates on triangulated graphs. A graph is triangulated (or chordal) iff every cycle composed of four or more vertices has a chord, which is an edge joining two vertices that are not adjacent in the cycle.

**Proposition 4.** *(Bliek & Sam-Haroud, 1999) Let $P$ be a binary CN $P$ with a triangulated constraint graph. $P$ is PPC-consistent iff $P$ is CPC-consistent.*

Enforcing path consistency simply means discarding path-inconsistent instantiations (i.e., recording new explicit nogoods of size two) since we know that the $PC$-closure, denoted $PC(P)$, of any CN $P$ exists (path consistency is well-behaved). To enforce path consistency it may be necessary to introduce new binary constraints, thus path consistency is not a conservative consistency. PPC and CPC differ in that only existing constraints are altered. As additional weak forms of path consistency, we find *directional* path consistency (Dechter & Pearl, 1988; Tsang, 1993) and *pivot* consistency (David, 1995), but they will not be discussed in this paper. Although these consistencies are attractive for controlling the practical inference effort in some situations, both consistencies require the introduction of a variable ordering, which restricts their applicability.

Finally, two domain-filtering consistencies related to path consistency have also been defined: *restricted path consistency* (RPC, Berlandier, 1995) and *max-restricted path consistency* (MaxRPC, Debruyne & Bessiere, 1997a). MaxRPC is strictly stronger than RPC and is defined as follows: a value $(x, a)$ of a CN $P$ is *max-restricted path consistent*, denoted *MaxRPC-consistent*, iff for every binary constraint $c_{xy}$ of $P$ involving $x$, there exists a locally consistent instantiation $\{(x, a), (y, b)\}$ of $\mathrm{scp}(c_{xy}) = \{x, y\}$ such that for every additional variable $z$ of $P$, there exists a value $c \in \mathrm{dom}(z)$ such that $\{(x, a), (z, c)\}$ and $\{(y, b), (z, c)\}$ are both locally consistent. A CN $P$ is MaxRPC-consistent iff every value of $P$ is MaxRPC-consistent.

## 3.3 Additional Second-Order Consistencies

In this section, we introduce second-order consistencies that are defined independently of path consistency. First, we recall 3-consistency (3C, Freuder, 1978).

**Definition 7** (3-consistency). *Let $P$ be a CN.*

- *An instantiation $\{(x, a), (y, b)\}$ on $P$ is* 3-consistent, *denoted* 3C-consistent, *iff there exists a value $c$ in the domain of every third variable $z$ of $P$ such that $\{(x, a), (y, b), (z, c)\}$ is locally consistent.*

- *$P$ is* 3-consistent *iff every locally consistent instantiation $\{(x, a), (y, b)\}$ on $P$ is 3-consistent.*

Note the difference with path consistency (see Definition 4): here, an instantiation of size 3 must be locally consistent (instead of two instantiations of size 2). It is known that 3C is equivalent to PC when no ternary constraint is present.

We now introduce dual consistency (Lecoutre et al., 2007a). Dual consistency, whose idea has initially been used by McGregor (1979), records inconsistent pairs of values identified by successive singleton checks. Just like singleton arc consistency, dual consistency is built on top of generalized arc consistency. Informally, a CN is dual-consistent iff each pair of values that is locally consistent is not detected inconsistent after assigning either of those two values and enforcing GAC. To simplify, we write $(x, a) \in P$ iff $(x, a)$ is a value of $P$, i.e., $x \in \mathrm{vars}(P) \wedge a \in \mathrm{dom}^P(x)$; when $P = \perp$, we consider that for every pair $(x, a)$, we have $(x, a) \notin P$.

**Definition 8** (Dual Consistency). *Let $P$ be a CN.*

- *An instantiation $\{(x, a), (y, b)\}$ on $P$ is* dual-consistent, *denoted* DC-consistent, *iff $(y, b) \in GAC(P|_{x=a})$ and $(x, a) \in GAC(P|_{y=b})$.*

- *$P$ is* DC-consistent *iff every locally consistent instantiation $\{(x, a), (y, b)\}$ on $P$ is DC-consistent.*

We may be interested by checks based on two simultaneous decisions (variable assignments): we obtain 2-singleton arc consistency (2SAC, Bessiere et al., 2005).

**Definition 9** (2-Singleton Arc Consistency). *Let $P$ be a CN.*

- *An instantiation $\{(x, a), (y, b)\}$ on $P$ is* 2-singleton arc-consistent, *denoted* 2SAC-consistent, *iff $GAC(P|_{\{x=a, y=b\}}) \neq \perp$.*

- $P$ is 2SAC-consistent *iff every locally consistent instantiation $\{(x,a),(y,b)\}$ on $P$ is 2SAC-consistent.*

3-consistency, dual consistency and 2-singleton arc consistency are second-order consistencies, from which conservative restrictions can be naturally derived as follows.

**Definition 10** (Conservative Second-Order Consistency). *Let $P$ be a CN, and $\phi$ be a consistency in $\{3C, DC, 2SAC\}$.*

- *An instantiation $\{(x,a),(y,b)\}$ on $P$ is* conservative $\phi$-consistent, *denoted* C$\phi$-consistent, *iff either $\not\exists c \in \text{cons}(P) \mid \text{scp}(c) = \{x,y\}$ or $\{(x,a),(y,b)\}$ is $\phi$-consistent.*

- *$P$ is C$\phi$-consistent iff every locally consistent instantiation $\{(x,a),(y,b)\}$ on $P$ is C$\phi$-consistent.*

Thus, we obtain three new second-order consistencies called conservative 3-consistency (C3C), conservative dual consistency (CDC, Lecoutre et al., 2007a) and conservative 2-singleton arc consistency (C2SAC). To illustrate the difference between a consistency $\phi$ and its conservative restriction $C\phi$, let us consider a CN $P$ such that $\text{vars}(P) = \{w,x,y,z\}$ and $\text{cons}(P) = \{c_{wx}, c_{wz}, c_{xyz}\}$, where subscripts indicate constraint scopes. $\phi$ reviews (locally consistent instantiations of) all of the six possible distinct pairs of variables whereas $C\phi$ reviews only the two pairs $(w,x)$ and $(w,z)$.

We shall also be interested in strong variants of second-order consistencies that additionally guarantee generalized arc consistency. For example, a binary CN is *strong path-consistent*, denoted *sPC-consistent*, iff it is both arc-consistent and path-consistent; a CN is *strong dual-consistent*, denoted *sDC-consistent* iff it is both GAC-consistent and DC-consistent. s3C, s2SAC, sPPC, sCPC, sCDC, sC3C, sC2SAC are defined similarly.

**Definition 11** (Strong Second-Order Consistency). *Let $\phi$ be a second-order consistency. A CN $P$ is strong $\phi$-consistent, denoted s$\phi$-consistent, iff $P$ is GAC+$\phi$-consistent, i.e., both GAC-consistent and $\phi$-consistent.*

A strong second-order consistency $\phi$ identifies both $\phi$-inconsistent values (nogoods of size one) and $\phi$-inconsistent pairs of values (nogoods of size two). Strictly speaking, this is not a second-order consistency but a "first+second" order consistency.

It is important to note that the closure of a CN can be computed for all second-order consistencies mentioned so far. All such consistencies can be proved to be stable, and consequently well-behaved.

**Proposition 5.** *PC, 3C, DC, 2SAC, PPC, CPC, C3C, CDC, C2SAC, as well as their strong variants, are consistencies that are well-behaved.*

*Sketch of proof.* Following Theorem 1, it is sufficient to show that all mentioned consistencies are stable (see Page 180). For each consistency $\phi$ among those mentioned in the proposition, when an instantiation $I$ is $\phi$-inconsistent on a CN $P$, necessarily if $I$ is not an explicit nogood in a CN $P'$ smaller than or equal to $P$, $I$ is $\phi$-inconsistent on $P'$. For example, suppose that $I = \{(x,a),(y,b)\}$ is DC-inconsistent on $P$. This means that $(y,b) \notin GAC(P|_{x=a})$ or $(x,a) \notin GAC(P|_{y=b})$. If $I \notin \widetilde{P'}$, we must then show that $I$ is DC-inconsistent on $P'$. But we necessarily have $(y,b) \notin GAC(P'|_{x=a}) \preceq GAC(P|_{x=a})$ or $(x,a) \notin GAC(P'|_{y=b}) \preceq GAC(P|_{y=b})$ because $P' \preceq P$. Hence, $I$ is DC-inconsistent on $P'$. $\qquad\square$

## 4. Relationships between Second-Order Consistencies

This section studies the qualitative relationships between the second-order consistencies that we have presented, namely path consistency, 3-consistency, dual consistency, 2-singleton arc consistency, and their conservative and strong variants. This section is composed of three parts (subsections). We start with relationships between basic second-order consistencies (PC, 3C, DC, 2SAC). Then, we focus on relationships including conservative restrictions (PPC, CPC, C3C, CDC, C2SAC). Finally, we finish with strong second-order consistencies (sPC, s3C, sDC, s2SAC, sPPC, sCPC, sC3C, sCDC, sC2SAC).

In our previous works (Lecoutre et al., 2007a, 2007b), our study was limited to binary CNs. In this paper, we generalize our results to CNs of any arity, although some results are given specifically for binary CNs or non-binary CNs. When no precision is given, the results hold for the set of all possible binary and non-binary CNs (i.e., CNs with constraints of arbitrary arity).

### 4.1 Results on Basic Second-Order Consistencies

We start with the strongest (basic) second-order consistency of this paper, namely 2SAC.

**Proposition 6.** *2SAC is strictly stronger than DC, and strictly stronger than 3C.*

*Proof.* Let $P$ be a CN and $I = \{(x, a), (y, b)\}$ be a locally consistent instantiation on $P$. On the one hand, if $I$ is DC-inconsistent then either $(y, b) \notin GAC(P|_{x=a})$ or $(x, a) \notin GAC(P|_{y=b})$, which necessarily entails $GAC(P|_{\{x=a,y=b\}}) = \bot$. Consequently, $I$ is 2SAC-inconsistent, and it follows that 2SAC is stronger than DC. On the other hand, if $I$ is 3C-inconsistent then $\exists z \in \text{vars}(P) \mid \forall c \in \text{dom}(z)$, $\{(x, a), (y, b), (z, c)\}$ is not locally consistent. In the CN $P' = GAC(P|_{\{x=a,y=b\}})$, we necessarily have $\text{dom}(z) = \emptyset$ (because $x$ and $y$ are assigned and GAC is enforced), and thus $P' = \bot$. This means that $I$ is 2SAC-inconsistent, and it follows that 2SAC is stronger than 3C. Strictness is proved by Figure 5 that shows a binary CN that is DC-consistent, 3C-consistent but not 2SAC-consistent.[6] $\square$

On binary CNs, DC is equivalent to PC. This could be predicted since McGregor proposed an AC-based algorithm to establish sPC (1979). We show this in 2 steps.

**Proposition 7.** *DC is strictly stronger than PC.*

*Proof.* Let $P$ be a CN and $I = \{(x, a), (y, b)\}$ be a locally consistent instantiation on $P$. If $I$ is path-inconsistent then $\exists z \in \text{vars}(P) \mid \forall c \in \text{dom}(z)$, $\{(x, a), (z, c)\}$ or $\{(y, b), (z, c)\}$ is not locally consistent (see Definition 4). In this case, we know that $(y, b) \notin GAC(P|_{x=a})$ since after enforcing GAC on $P|_{x=a}$, every value $c$ remaining in $\text{dom}(z)$ is such that $\{(x, a), (z, c)\}$ is consistent. Necessarily, by hypothesis, all these remaining values are incompatible with $(y, b)$, thus $b$ is removed from $\text{dom}(y)$ when enforcing GAC. Hence $I$ is dual-inconsistent, and it follows that DC is stronger than PC. Strictness is proved by Figure 6 that shows a non-binary CN that is PC-consistent but not DC-consistent. $\square$

**Proposition 8.** *On binary CNs, DC is equivalent to PC.*

---

6. This result, as well as those of Figures 9 and 12, has been computer-checked.

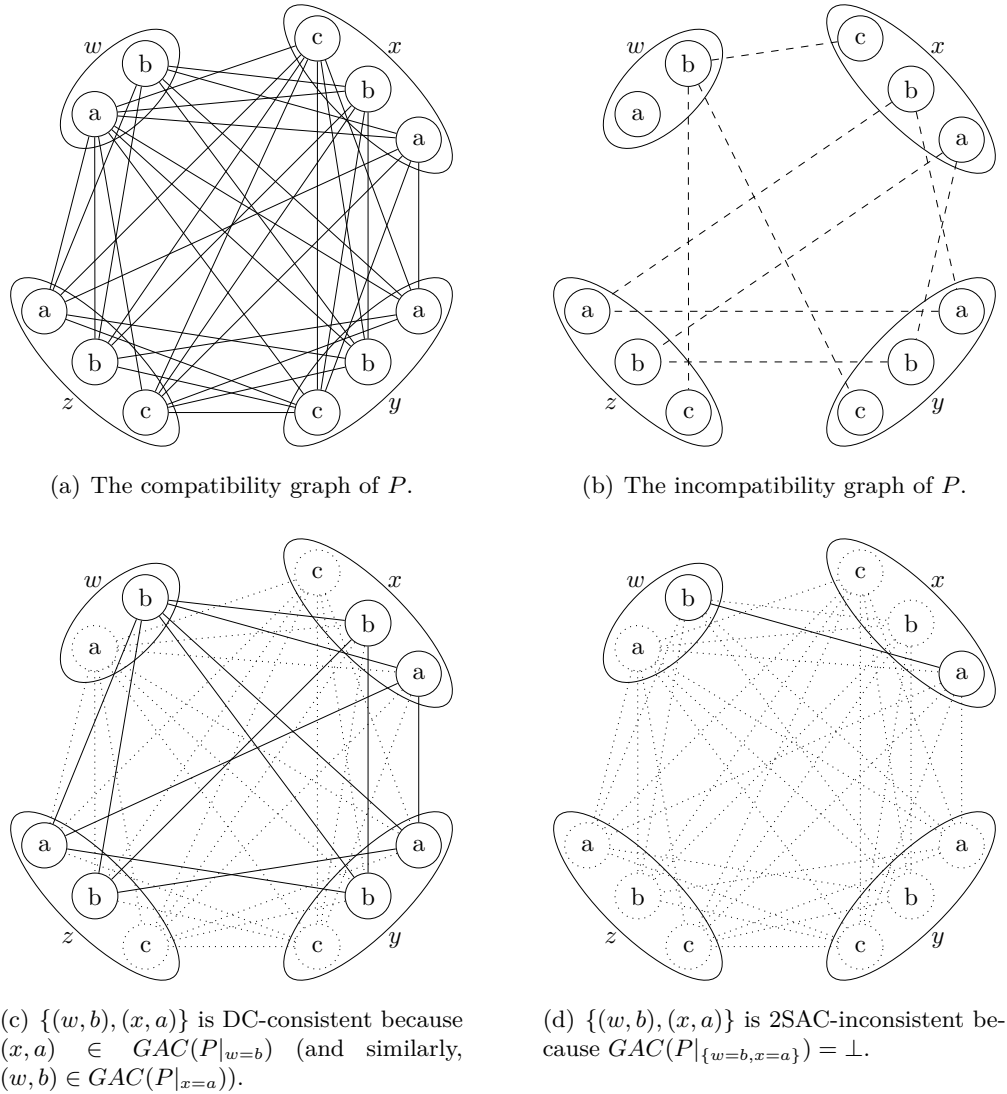

(a) The compatibility graph of $P$.

(b) The incompatibility graph of $P$.

(c) $\{(w, b), (x, a)\}$ is DC-consistent because $(x, a) \in GAC(P_{|w=b})$ (and similarly, $(w, b) \in GAC(P_{|x=a})$).

(d) $\{(w, b), (x, a)\}$ is 2SAC-inconsistent because $GAC(P_{|\{w=b, x=a\}}) = \bot$.

Figure 5: A CN $P$ that is sDC-consistent (and s3C-consistent) but not C2SAC-consistent. Dotted circles and lines correspond to deleted values and tuples.

*Proof.* From Proposition 7, we know that DC is stronger than PC. Now, we show that, on binary CNs, PC is stronger than DC, therefore we can conclude that DC and PC are equivalent. Let $P$ be a binary CN and $I = \{(x, a), (y, b)\}$ be a locally consistent instantiation on $P$. If $I$ is dual-inconsistent then $(y, b) \notin AC(P_{|x=a})$, or symmetrically $(x, a) \notin AC(P_{|y=b})$. We consider the first case.

Let us consider a filtering procedure F that iteratively removes (in any order) the values of $P_{|x=a}$ that are successively found to be arc-inconsistent up to a fixpoint. Such a procedure is guaranteed to compute $AC(P_{|x=a})$ (cf. Apt, 2003; Lecoutre, 2009). Let $H(k)$ be the following induction hypothesis: if $(z, c)$ is one of the $k$ first values removed by F,

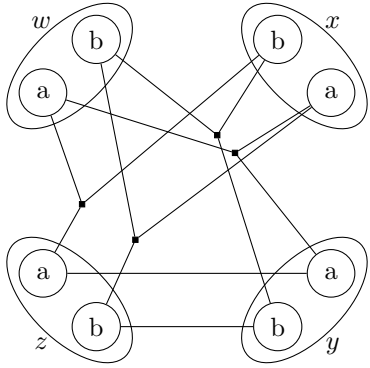

Figure 6: A CN $P$ with two ternary constraints $c_{wxy}$ and $c_{wxz}$ such that $\mathrm{rel}(c_{wxy}) = \{(a,a,a), (b,b,b)\}$ and $\mathrm{rel}(c_{wxz}) = \{(a,b,a), (b,a,b)\}$. $P$ also involves a binary constraint $c_{yz}$. $P$ is sPC-consistent but not CDC-consistent. Indeed, $\{(y,a),(z,a)\}$ is DC-inconsistent since $GAC(P|_{y=a}) = \bot$.

then $\{(x,a),(z,c)\} \in \widetilde{P'}$ with $P' = PC(P)$, i.e., $\{(x,a),(z,c)\}$ is either initially locally inconsistent or identified as path-inconsistent (possibly after propagation).

We show that $H(1)$ holds. If $(z,c)$ is the first value removed by F, this means that $(z,c)$ has no support on a binary constraint involving $z$ and a second variable $w$. If $\{(x,a),(z,c)\}$ is locally inconsistent, then $H(1)$ holds trivially. Otherwise, necessarily $w \neq x$ (because this would mean that $(z,c)$ is not compatible with $(x,a)$ since $a$ has been assigned to $x$, so $\{(x,a),(z,c)\}$ is initially locally inconsistent). Therefore $\{(x,a),(z,c)\}$ clearly has no support on the path $\langle x,w,z \rangle$ and is thus path-inconsistent.

We now assume that $H(k)$ is true and show that $H(k+1)$ holds. If $(z,c)$ is the $k+1$th value removed by F, this means that this removal involves a constraint binding $z$ with another variable $w$. The value $(z,c)$ has no support on this constraint, thus every value in $\mathrm{dom}(w)$ initially supporting $(z,c)$, if any, is one of the $k$ first values removed by F. By hypothesis this means that for any such value $b$, $\{(x,a),(w,b)\} \in \widetilde{P'}$ with $P' = PC(P)$. In any case, we can now deduce that $\{(x,a),(z,c)\} \in \widetilde{P'}$ and, as a special case, we can identify $I$ as path-inconsistent. Consequently, every locally consistent instantiation on $P$ that is in $\widetilde{P''}$ with $P'' = DC(P)$ is also in $\widetilde{P'}$ with $P' = PC(P)$. We deduce that $PC(P) \preceq DC(P)$ and also from Proposition 1 that PC is stronger than DC on binary CNs. $\qquad\square$

Now, we consider 3-consistency. On binary CNs, it is well-known that 3-consistency is equivalent to path consistency. So, 3C is also equivalent to DC (since DC is equivalent to PC). On non-binary CNs, we have the following relationships with PC and DC.

**Proposition 9.** *On non-binary CNs, 3C is incomparable with DC, and strictly stronger than PC.*

*Proof.* On non-binary CNs, 3C is strictly stronger than PC (e.g., see Dechter, 2003, p. 69) because 3C also checks ternary constraints. When comparing 3C with DC, it appears that 3C cannot be stronger than DC because a CN composed of only one quaternary constraint that is not GAC-consistent is necessarily 3-consistent (since there are no binary and ternary constraints) and not DC-consistent. On the other hand, DC cannot be stronger than 3C because Figure 7 shows a non-binary CN that is DC-consistent but not 3-consistent (there is no way of extending $\{(x,a),(y,a)\}$ on $z$). Hence, 3C and DC are incomparable. □

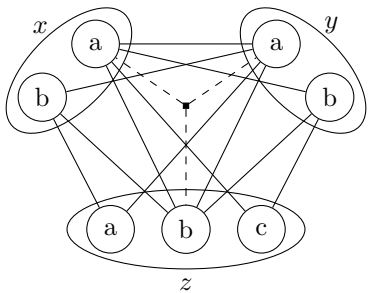

Figure 7: A CN $P$ that is sDC-consistent but not C3C-consistent (because $\{(x,a),(y,a)\}$ is not 3-consistent). The dashed hyperedge corresponds to the nogood $\{(x,a),(y,a),(z,b)\}$, i.e., a ternary constraint $c_{xyz}$ only forbidding the tuple $(a,a,b)$.

## 4.2 Results on Conservative Second-Order Consistencies

Now, we consider conservative variants of the basic second-order consistencies. A first immediate result is that conservative consistencies are made strictly weaker than their unrestricted forms. However, it is worthwhile to mention that (strong) PPC was shown equivalent to (strong) PC on binary convex CNs with triangulated constraint graphs (Bliek & Sam-Haroud, 1999).

**Proposition 10.** *2SAC, DC, 3C and PC are respectively strictly stronger than C2SAC, CDC, C3C and PPC+CPC.*

*Proof.* By definition, conservative consistencies are weaker. Strictness is proved by Figure 8 that shows a binary CN which is C2SAC-consistent (and CDC-consistent, C3C-consistent, PPC-consistent, CPC-consistent) but not PC-consistent (and not 3C-consistent, not DC-consistent and not 2SAC-consistent). □

**Proposition 11.** *PC, DC and 3C are incomparable with C2SAC.*

*Proof.* On the one hand, PC cannot be stronger than C2SAC since Figure 5 shows a binary CN that is PC-consistent but not C2SAC-consistent. On the other hand, C2SAC cannot be stronger than PC since Figure 8 shows a binary CN that is C2SAC-consistent but not PC-consistent. We conclude that PC and C2SAC are incomparable. This is also valid for DC and 3C since only binary CNs are mentioned in this proof (and PC=DC=3C on binary CNs). □

**Proposition 12.** *C2SAC is strictly stronger than CDC, and strictly stronger than C3C.*

*Proof.* The proof is similar to that of Proposition 6, considering initially a locally consistent instantiation $I = \{(x,a), (y,b)\}$ which is CDC-inconsistent (and next C3C-inconsistent) and such that $x$ is linked to $y$ by a constraint. $\square$

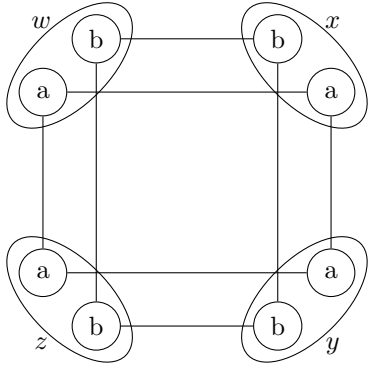

Figure 8: A CN (no constraint binds $w$ with $y$ and $x$ with $z$) that is sC2SAC-consistent (and CDC+C3C+PPC+CPC-consistent), but not PC-consistent (and not 2SAC-consistent). For example, $\{(x,a), (z,b)\}$ is not path-consistent.

**Proposition 13.** *CDC is strictly stronger than PPC.*

*Proof.* Assume that a CN $P$ is CDC-consistent and consider a closed graph-path $\langle x_1, \ldots, x_p \rangle$ of $P$. For every locally consistent instantiation $\{(x_1, a_1), (x_p, a_p)\}$ on $P$, $(x_p, a_p) \in P'$ with $P' = GAC(P|_{x_1 = a_1})$ since $P$ is CDC-consistent. It also implies $P' \neq \perp$. Therefore, in the context of $P'$, there exists at least one value in each domain and since $P'$ is generalized arc-consistent, there is clearly a value $(x_{p-1}, a_{p-1})$ of $P'$ compatible with $(x_p, a_p)$, a value $(x_{p-2}, a_{p-2})$ of $P'$ compatible with $(x_{p-1}, a_{p-1})$, ..., and a value $(x_1, a'_1)$ of $P'$ compatible with $(x_2, a_2)$. Because $\text{dom}^{P'}(x_1) = \{a_1\}$, we have $a'_1 = a_1$, thus the locally consistent instantiation $\{(x_1, a_1), (x_p, a_p)\}$ is consistent on the closed graph-path $\langle x_1, \ldots, x_p \rangle$ of $P$. Hence $P$ is PPC-consistent, thus CDC is stronger than PPC.

The fact that CDC is strictly stronger than PPC is shown by the CN $P$ depicted in Figure 9. In Figure 9(c), $P$ is shown to be CDC-inconsistent because the locally consistent instantiation $\{(x,a), (y,b)\}$ is dual-inconsistent: $(y,b) \notin AC(P|_{x=a})$. In Figure 9(d), $P$ is shown to be CPC-consistent because, for example, the locally consistent instantiation $\{(x,a), (y,b)\}$ is consistent on all 2-length graph-paths linking $x$ to $y$, namely, $\langle x, z, y \rangle$ and $\langle x, w, y \rangle$. Here, the constraint graph is triangulated, which means that CPC is equivalent to PPC. Hence we can deduce our result. $\square$

**Proposition 14.** *On non-binary CNs, CDC is incomparable with C3C.*

*Proof.* CDC cannot be stronger than C3C since Figure 7 shows a non-binary CN that is CDC-consistent but not C3C-consistent. Now, consider the CN of Figure 9 extended with a single ternary constraint involving new variables. This CN remains C3C-consistent (because

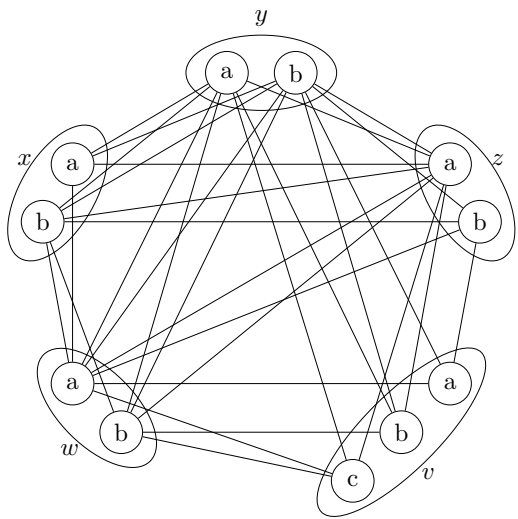

(a) The compatibility graph of $P$ (no constraint binds $x$ with $v$).

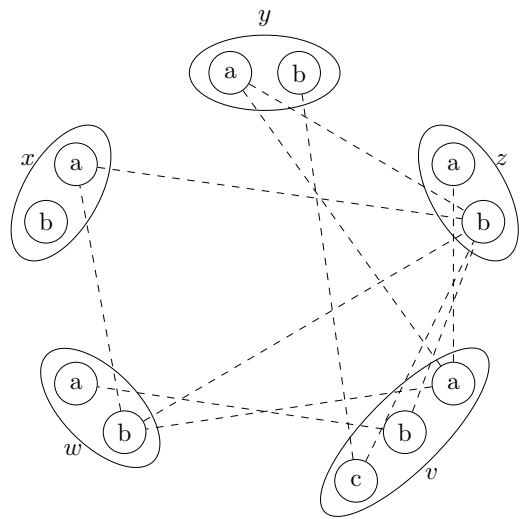

(b) The incompatibility graph of $P$.

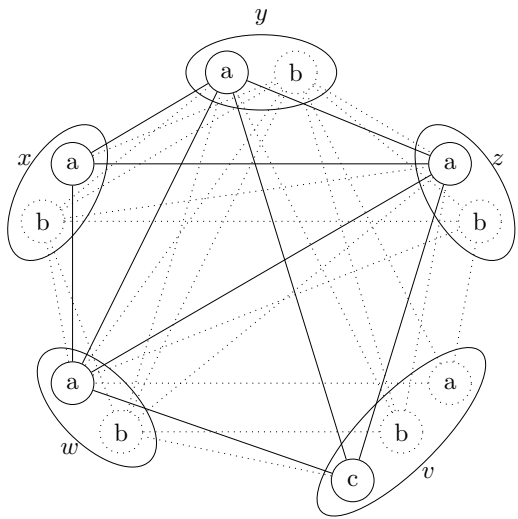

(c) $P$ is *not* CDC-consistent. We can see that $(y, b) \notin AC(P|_{x=a})$. Thus, the locally consistent instantiation $\{(x, a), (y, b)\}$ is dual-inconsistent.

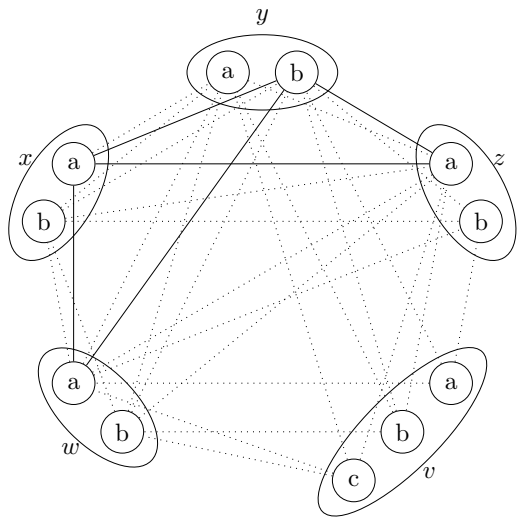

(d) $P$ is sCPC-consistent (and hence sPPC-consistent since $P$ is triangulated). Any (closed) 2-length graph-path of $P$ linking $x$ to $y$ is consistent. This is shown here for $\{(x, a), (y, b)\}$.

Figure 9: Example of a binary CN $P$ that is sPPC-consistent (and sC3C-consistent) but not CDC-consistent.

there is no binary constraint between these new variables), but is still not CDC-consistent. Hence, C3C cannot be stronger than CDC, and CDC and C3C are incomparable. $\square$

**Proposition 15.** *On binary CNs, PPC is strictly stronger than C3C. On non-binary CNs, PPC is incomparable with C3C.*

*Proof.* 1) Let $P$ be a binary CN that is PPC-consistent (and different from $\perp$ – this is a very weak restriction). We consider a locally consistent instantiation $\{(x,a),(y,b)\}$ on $P$ (such that there is a binary constraint involving $x$ and $y$) and show that for every third variable $z$ of $P$, the following property $Pr(z)$ holds: $\exists c \in \text{dom}(z)$ such that $\{(x,a),(z,c)\}$ and $\{(y,b),(z,c)\}$ are both locally consistent instantiations, which is equivalent to "$\{(x,a),(y,b),(z,c)\}$ is locally consistent" since $P$ is binary. If $Pr$ holds, then $\{(x,a),(y,b)\}$ is C3C-consistent. For each variable $z$, 3 cases must be considered, depending of the existence of the constraints $c_{xz}$, between $x$ and $z$, and $c_{yz}$, between $y$ and $z$.

(a) Both constraints exist: thus, there exists a graph-path $\langle x, z, y \rangle$ and as this path is consistent by hypothesis, necessarily the property $Pr(z)$ holds.

(b) Neither constraint exist: $P \neq \perp$ implies $\text{dom}(z) \neq \emptyset$, thus $Pr(z)$ holds because $c_{xz}$ and $c_{yz}$ are implicit and universal.

(c) Only the constraint $c_{xz}$ exists (similarly, only the constraint $c_{yz}$ exists). Consider the graph-path $\langle x, z, x, y \rangle$. By hypothesis, this graph-path is consistent. Hence, there exists a value $c$ in $dom(z)$ such that $\{(x,a),(z,c)\}$ is locally consistent. We also know that $\{(y,b),(z,c)\}$ is locally consistent because there is no constraint between $y$ and $z$. We conclude that $Pr(z)$ holds and PPC is stronger than C3C on binary CNs.

Figure 10 proves strictness by showing a CN that is C3C-consistent but not PPC-consistent.

2) For non-binary CNs, Figure 7 shows that PPC cannot be stronger than C3C: the CN is PPC-consistent but not C3C-consistent. Now, consider the CN of Figure 10 extended with a single ternary constraint involving new variables. This CN remains C3C-consistent (because there is no binary constraint between these new variables), but is still not PPC-consistent. Hence, C3C cannot be stronger than PPC, and PPC and C3C are incomparable. $\square$

**Proposition 16.** *C3C is strictly stronger than CPC.*

*Proof.* Let $P$ be a CN that is C3C-consistent. Let $\langle x, z, y \rangle$ be a closed 2-length graph-path of $P$ and $\{(x,a),(y,b)\}$ be a locally consistent instantiation on $P$. Because $P$ is C3C-consistent, we know that there exists a value $c$ in $\text{dom}(z)$ such that $\{(x,a),(y,b),(z,c)\}$ is locally consistent. Consequently, there exists a value $c$ in $\text{dom}(z)$ such that $\{(x,a),(z,c)\}$ and $\{(y,b),(z,c)\}$ are both locally consistent. We deduce that the path $\langle x, z, y \rangle$ is consistent, thus $P$ is CPC-consistent and C3C is stronger than CPC. Figure 11 proves strictness by showing a binary CN that is CPC-consistent (there is no 3-clique) but not C3C-consistent. $\square$

**Proposition 17.** *PPC is strictly stronger than CPC.*

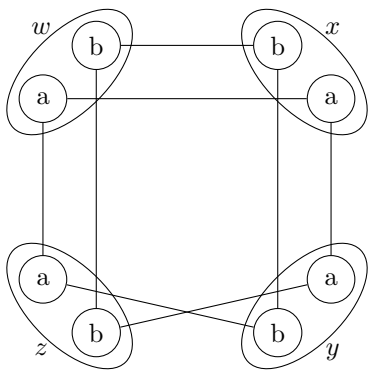

Figure 10: A CN (no constraint binds $w$ with $y$ and $x$ with $z$) that is sC3C-consistent (and sCPC-consistent, and BiSAC-consistent) but not PPC-consistent. For example, $\{(x, a), (w, a)\}$ is not PPC-consistent.

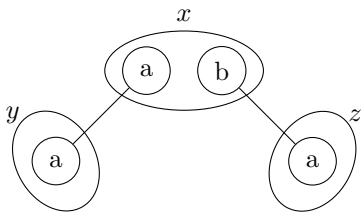

Figure 11: A binary CN $P$ with two constraints (no constraint exists between $y$ and $z$). $P$ is CPC-consistent but not C3C-consistent.

*Proof.* PPC is stronger than CPC by definition. Moreover, the binary CN in Figure 10 is CPC-consistent but not PPC-consistent. Because there is no 3-clique in its constraint graph, this CN is trivially CPC-consistent. □

**Proposition 18.** *On non-binary CNs, PC is incomparable with CDC.*

*Proof.* On the one hand, consider the CN of Figure 8 extended with a single ternary GAC-consistent constraint involving new variables. Because the additionnal constraint is GAC-consistent, the CN remains CDC-consistent. However, it is not PC-consistent. We deduce that CDC cannot be stronger than PC (on non-binary CNS). On the other hand, Figure 6 proves that PC cannot be stronger than CDC: $P$ is PC-consistent (because there is only one binary constraint) but not CDC-consistent ($\{(y, a), (z, a)\}$ is CDC-inconsistent). We conclude that on non-binary CNs, PC and CDC are incomparable. □

## 4.3 Results on Strong Second-Order Consistencies

Before studying the relationships existing between strong variants of second-order consistencies, we observe that, in the binary case, enforcing AC on a path-consistent CN is sufficient to obtain a strong path-consistent CN. This well-known fact is also true in the general case for DC, CDC, 2SAC and C2SAC. We define $\phi \circ \psi(P)$ as being $\phi(\psi(P))$.

**Proposition 19.** *For any binary CN $P$, we have $AC \circ PC(P) = sPC(P)$.*

*Proof.* In $PC(P)$, every locally consistent instantiation $\{(x, a), (y, b)\}$ has a support on every third variable $z$. Hence, every value of $PC(P)$ in a locally consistent instantiation is arc-consistent. Consequently, when AC is enforced on $PC(P)$, no value present in a locally consistent instantiation (of size 2) can be removed, and PC is preserved. $\square$

**Proposition 20.** *For any CN $P$, we have $GAC \circ DC(P) = sDC(P)$, $GAC \circ CDC(P) = sCDC(P)$, $GAC \circ 2SAC(P) = s2SAC(P)$ and $GAC \circ C2SAC(P) = sC2SAC(P)$.*

*Proof.* Let $P' = DC(P)$ and $P'' = GAC(P')$. For any singleton check $GAC(P''|_{x=a})$ on $P''$, we have $GAC(P''|_{x=a}) = GAC(GAC(P')|_{x=a}) = GAC(P'|_{x=a})$. This means that the result of the singleton check for $(x, a)$ on $P''$ is the same as the result of the singleton check for $(x, a)$ on $P'$. Since $P'$ is DC-consistent, we can deduce that $DC(P'') = P''$. $P''$ being both GAC-consistent and DC-consistent, we have $P'' = GAC \circ DC(P) = sDC(P)$. A similar proof holds for *CDC*, *2SAC* and *C2SAC*. $\square$

It has been shown that the schema of previous propositions does not hold for CPC and PPC (Lecoutre, 2009). For example, for some binary CNs $P$, we have $AC \circ CPC(P) \neq sCPC(P)$. Unsurprisingly, some relationships are preserved when strong variants are considered.

**Proposition 21.** *Let $\phi$ and $\psi$ be two second-order consistencies. If $\phi$ is stronger than $\psi$ then $s\phi$ is stronger than $s\psi$.*

**Proposition 22.** *We have:*

(a) *s2SAC is strictly stronger than sDC, and strictly stronger than s3C.*

(b) *sDC is strictly stronger than sPC.*

(c) *s3C is strictly stronger than sPC.*

(d) *s2SAC, sDC, s3C, and sPC are respectively strictly stronger than sC2SAC, sCDC, sC3C, and sPPC+sCPC.*

(e) *sC2SAC is strictly stronger than sCDC, and strictly stronger than sC3C.*

(f) *sCDC is strictly stronger than sPPC.*

(g) *sPPC is strictly stronger than sCPC.*

*Proof.* All illustrative CNs introduced previously are GAC-consistent (except for Figure 11), thus using Proposition 21, it suffices to consider: (a) Proposition 6 with Figure 5 for strictness, (b) Proposition 7 with Figure 6 for strictness, (c) Proposition 9 with Figure 7 for strictness, (d) Proposition 10 with Figure 8 for strictness, (e) Proposition 12 with Figure 5 for strictness, (f) Proposition 13 with Figure 9 for strictness, (g) Proposition 17 with Figure 10 for strictness. $\square$

The following result indicates that C3C and CPC are quite close properties. For arc-consistent CNs, they are equivalent.

**Proposition 23.** *On binary CNs, sC3C is equivalent to sCPC.*

*Proof.* From Propositions 16 and 21, we know that sC3C is stronger than sCPC. Now, we show that, on binary CNs, sCPC is stronger than sC3C. The proof is similar to that of Proposition 15 by considering a binary CN that is initially sCPC-consistent. Only case (c) in the demonstration differs:

(c) Only the constraint $c_{xz}$ exists (similarly, only the constraint $c_{yz}$ exists): as $P$ is arc-consistent, there exists a value in $\text{dom}(z)$ that is compatible with $(x, a)$. Because there is an implicit universal constraint between $z$ and $y$, this value is also compatible with $(y, b)$. Hence, $Pr(z)$ holds. □

**Proposition 24.** *sPC, sDC and s3C are incomparable with sC2SAC.*

*Proof.* In the proof of Proposition 11, the CNs of Figures 5 and 8 are GAC-consistent. □

**Proposition 25.** *On non-binary CNs, sPC is incomparable with sCDC.*

*Proof.* In the proof of Proposition 18, the CNs of Figures 8 and 6 are GAC-consistent. □

Finally, we conclude this section by establishing some connections with SAC.

**Proposition 26.** *sDC is strictly stronger than SAC+CDC*

*Proof.* Let $P$ be a CN that is sDC-consistent. Assume that a value $(x, a)$ of $P$ is SAC-inconsistent. This means that $P' = GAC(P|_{x=a}) = \bot$, and for every value $(y, b)$, we have $(y, b) \notin P'$ (recall that no value belongs to $\bot$). As $P$ is DC-consistent by hypothesis, these nogoods are recorded in $P$ meaning that for every binary constraint $c_{xy}$ forbidding any tuple involving $(x, a)$. We deduce that $(x, a)$ is GAC-inconsistent, which contradicts our hypothesis ($P$ sDC-consistent), and shows that sDC is stronger than SAC. As we know that DC is strictly stronger than CDC, we deduce that sDC is stronger than SAC+CDC. To prove strictness, it suffices to build a CN that is SAC-consistent, CDC-consistent but not DC-consistent (e.g., see Figure 8). □

**Proposition 27.** *On binary CNs, sCDC is strictly stronger than SAC.*

*Proof.* Let $P$ be a binary CN that is sCDC-consistent. Assume that a value $(x, a)$ of $P$ is SAC-inconsistent. This means that $AC(P|_{x=a}) = \bot$. As $P$ is AC-consistent (since $P$ is sCDC-consistent by hypothesis), necessarily $x$ is involved in (at least) a binary constraint $c$ (otherwise no propagation is possible to deduce $AC(P|_{x=a}) = \bot$). Consequently, there is no tuple allowed by $c$ involving $(x, a)$ since $P$ is CDC-consistent (because when $P' = \bot$, for every value $(y, b)$, we consider $(y, b) \notin P'$). We deduce that $(x, a)$ is AC-inconsistent. This contradiction shows that sCDC is stronger than SAC. To prove strictness, it suffices to observe that sCDC reasons with both inconsistent values and pairs of values. □

**Proposition 28.** *On binary CNs, SAC+CDC is equivalent to sCDC. On non-binary CNs, SAC+CDC is strictly stronger than sCDC.*

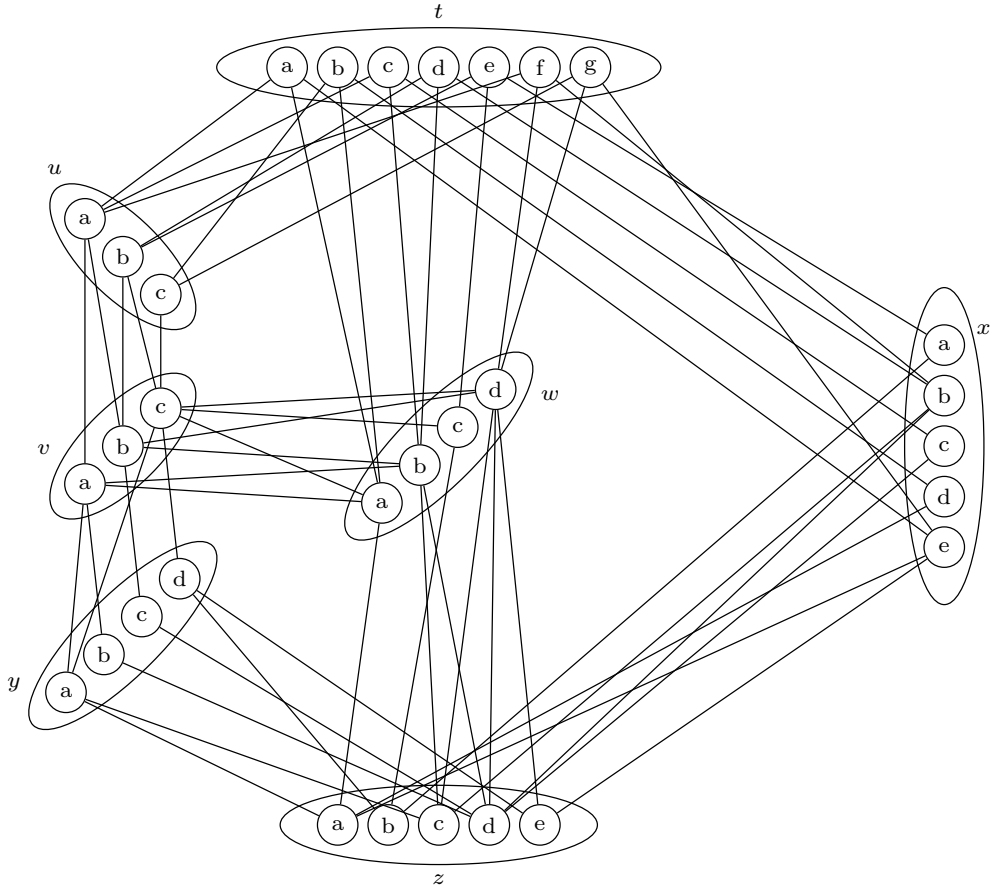

Figure 12: A CN that is sCDC-consistent but not BiSAC-consistent: $(z, c)$ is not BiSAC-consistent, as this value appears in none of the singleton tests $AC(P|_{v=a})$, $AC(P|_{v=b})$ and $AC(P|_{v=c})$.

*Proof.* Clearly, SAC+CDC is stronger than sCDC since SAC is stronger than GAC (and sCDC is GAC+CDC). On the other hand, sCDC is trivially stronger than CDC and we know from Proposition 27 that sCDC is stronger than SAC on binary CNs. We deduce that, on binary CNs, sCDC is stronger than SAC+CDC, and that SAC+CDC is equivalent to sCDC. For non-binary CNs, to show strictness, let us consider the non-binary CN depicted in Figure 6, but with the binary constraint $c_{yz}$ eliminated. The new obtained CN is GAC-consistent, CDC-consistent (since there are no more binary constraints), and thus sCDC-consistent, but not SAC-consistent because $GAC(P|_{y=a}) = \bot$.  □

**Proposition 29.** *On binary CNs, sCDC is incomparable with BiSAC.*

*Proof.* On the one hand, BiSAC cannot be stronger than sCDC since Figure 9 shows a binary CN that is BiSAC-consistent (note that every value belongs to at least one solution) but not CDC-consistent. On the other hand, sCDC cannot be stronger than BiSAC since Figure 12

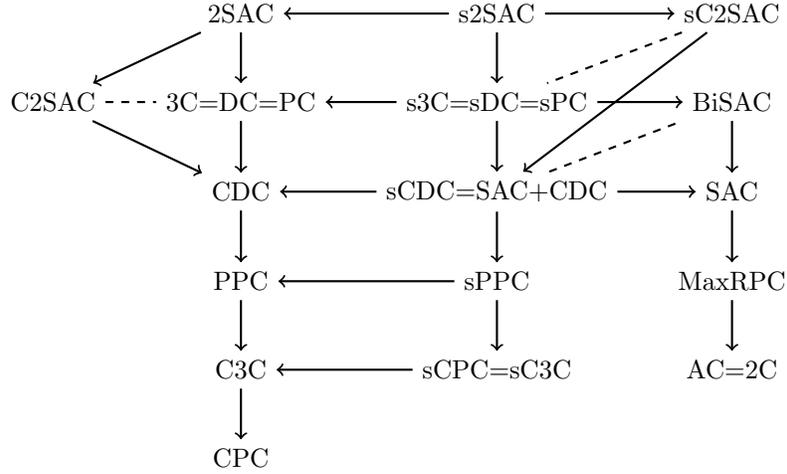

(a) Consistencies restricted to binary CNs.

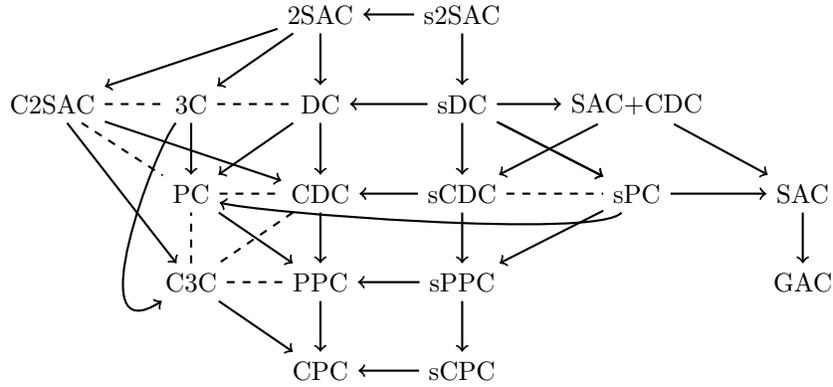

(b) Consistencies for CNs with constraints of arbitrary arity.

$\longrightarrow$ strictly stronger     - - - incomparable     = equivalent

Figure 13: Summary of the relationships between consistencies.

shows a binary CN that is sCDC-consistent but not BiSAC-consistent. We conclude that sCDC and BiSAC are incomparable. □

Some results given in the general case (i.e., for CNs with constraints of arbitrary arity) also hold when only binary CNs are considered. This is the case for Propositions 6, 10, 11, 12, 13, 16, 17, 22 (except for cases (b) and (c)), 24 and 26 because binary CNs are used in their proofs. Figure 13 shows the relationships between (strong) second-order consistencies introduced in this paper, with a focus on binary CNs in Figure 13(a). In Figure 13(b), for the sake of clarity, s3C, sC3C, and sC2SAC are not inserted.

## 5. An Algorithm to Enforce s(C)DC

In this section, we present a general algorithm to enforce strong (conservative) dual consistency. This algorithm is valid for both binary and non-binary CNs. This algorithm is called sCDC1 when it is used to enforce sCDC, and sDC1 when it is used to enforce sDC. Actually, on non-binary CNs, our sCDC1 algorithm enforces SAC+CDC, which is strictly stronger than sCDC. This extra strength comes for free from the exploitation of the singleton checks used to enforce CDC.

---

**Algorithm 1**: sCDC1/sDC1($P$): Boolean

**Input/Output**: a CN $P$; sCDC (SAC+CDC) or sDC is enforced on $P$
**Result**: true iff $P$ is strong (conservative) dual-consistent

1  $P \leftarrow GAC(P)$                    // GAC is initially enforced
2  **if** $P = \bot$ **then return false**
3  $x \leftarrow \mathsf{first}(\mathrm{vars}(P))$
4  $marker \leftarrow x$
5  **repeat**
6    | **if** $reviseVariable(P, x)$ **then**
7    |   | $P \leftarrow GAC(P)$                // GAC is maintained
8    |   | **if** $P = \bot$ **then return false**
9    |   | $marker \leftarrow x$
10   | $x \leftarrow \mathsf{nextCircular}(x, \mathrm{vars}(P))$
11 **until** $x = marker$
12 **return true**

---

Algorithm 1 establishes strong (conservative) dual consistency on a given CN $P$. The $\mathsf{learn}^{xxx}$ function called at Line 9 of Algorithm 2, which is used by Algorithm 1, is specialized either to $\mathsf{learn}^{part}$ (Algorithm 3) to enforce sCDC (SAC+CDC on non-binary CNs – we omit this precision from now on) or to $\mathsf{learn}^{full}$ (Algorithm 4) to enforce sDC on $P$. Basically, Algorithm 1 performs successive singleton checks until a fixed point is reached, and returns **true** iff $P$ is strong (conservative) dual-consistent, i.e., iff $sCDC(P) \neq \bot$ (with $\mathsf{learn}^{part}$) or $sDC(P) \neq \bot$ (with $\mathsf{learn}^{full}$). GAC is enforced at line 1, and then a variable is considered at each turn of the main loop to establish the consistency. $\mathsf{first}(\mathrm{vars}(P))$ is the first variable of $P$ in the lexicographical order, and $\mathsf{nextCircular}(x, \mathrm{vars}(P))$ is the variable of $P$ right after $x$ if any, or $\mathsf{first}(\mathrm{vars}(P))$ otherwise. These two functions allow circular iteration over the variables of $P$. For example, if $\mathrm{vars}(P) = \{x, y, z\}$, then the iteration has the form $x$, $y$, $z$, $x$, $y$, $z \ldots$ Of course, it is possible to control the order of variables from one iteration to the next using some heuristic.

The $\mathsf{reviseVariable}$ function (Algorithm 2) revises the given variable $x$ by means of strong (conservative) dual consistency, i.e., explores all possible inferences with respect to $x$ by performing singleton checks on values in $\mathrm{dom}(x)$. To achieve this, GAC is enforced on $P|_{x=a}$ for each value $a$ in the domain of $x$ (Line 3). If $a$ is SAC-inconsistent, then $a$ is removed from the domain of $x$ (Line 5). Otherwise (Lines 7 to 10), for every variable $y \neq x$ of $P$ with at least one value deleted by constraint propagation, we try to learn nogoods by

---

**Algorithm 2**: reviseVariable($P$,$x$): Boolean

---

**1**   *effective* $\leftarrow$ **false**

**2**   **foreach** *value* $a \in \text{dom}^P(x)$ **do**

**3**     $P' \leftarrow GAC(P|_{x=a})$             // singleton check on $(x,a)$

**4**     **if** $P' = \perp$ **then**

**5**       remove $a$ from $\text{dom}^P(x)$        // SAC-inconsistent value

**6**       *effective* $\leftarrow$ **true**

**7**     **else**

**8**       **foreach** *variable* $y$ *in* $\text{vars}(P) \mid y \neq x \wedge \text{dom}^P(y) \neq \text{dom}^{P'}(y)$ **do**

**9**         **if** *learn*$^{xxx}(P,(x,a),y,\text{dom}^P(y) \setminus \text{dom}^{P'}(y))$ **then**

**10**           *effective* $\leftarrow$ **true**

**11**   **return** *effective*

---

**Algorithm 3**: learn$^{part}(P, (x,a), y, Deleted)$: Boolean

---

**1**   **if** $\exists c_{xy} \in \text{cons}(P) \mid \text{scp}(c_{xy}) = \{x,y\}$ **then**

**2**     *conflicts* $\leftarrow \emptyset$

**3**     **foreach** $b \in Deleted \mid (a,b) \in \text{rel}^P(c_{xy})$ **do**

**4**       *conflicts* $\leftarrow$ *conflicts* $\cup \{(a,b)\}$       // CDC-inconsistent pair

**5**     **if** *conflicts* $\neq \emptyset$ **then**

**6**       $\text{rel}^P(c_{xy}) \leftarrow \text{rel}^P(c_{xy}) \setminus$ *conflicts*

**7**       **return true**

**8**   **return false**

---

**Algorithm 4**: learn$^{full}(P, (x,a), y, Deleted)$: Boolean

---

**1**   **if** $\exists c_{xy} \in \text{cons}(P) \mid \text{scp}(c_{xy}) = \{x,y\}$ **then**

**2**     *conflicts* $\leftarrow \emptyset$

**3**     **foreach** $b \in Deleted \mid (a,b) \in \text{rel}^P(c_{xy})$ **do**

**4**       *conflicts* $\leftarrow$ *conflicts* $\cup \{(a,b)\}$       // CDC-inconsistent pair

**5**     **if** *conflicts* $\neq \emptyset$ **then**

**6**       $\text{rel}^P(c_{xy}) \leftarrow \text{rel}^P(c_{xy}) \setminus$ *conflicts*

**7**       **return true**

**8**     **else**

**9**       *conflicts* $\leftarrow \{(a,b) \mid b \in Deleted\}$       // DC-inconsistent pairs

**10**       Let $c_{xy}$ be a new constraint such that:
          •   $\text{scp}(c_{xy}) = \{x,y\}$
          •   $\text{rel}(c_{xy}) = (\text{dom}^{init}(x) \times \text{dom}^{init}(y)) \setminus$ *conflicts*

**11**       $\text{cons}(P) \leftarrow \text{cons}(P) \cup \{c_{xy}\}$

**12**       **return true**

**13**   **return false**

---

means of functions $\mathsf{learn}^{part}$ or $\mathsf{learn}^{full}$; the set of values of $y$ deleted by propagation is passed as last parameter (in practice, using a stack to handle domains and a trailing mechanism, there is no need to explicitly compute the set of deleted values). The revision is effective for $x$ if a value or a tuple is deleted (possibly after inserting a new constraint). The Boolean variable *effective* is introduced to track revision effectiveness. When the revision of $x$ is effective, reviseVariable returns **true** at Line 6 of Algorithm 1, and GAC is re-established (Line 7). Any domain or relation wipe-out is detected at Line 8. A marker, initialized with the first variable of vars($P$) (Line 4) and updated whenever inferences are performed (Line 9), manages termination.

Both Algorithms 3 and 4 discard identified binary nogoods that correspond to CDC-inconsistent pairs of values if the constraint $c_{xy}$ exists: every tuple $(a, b)$ such that $b$ is a deleted value (i.e., $b$ for $y$ is present in $P$ but not in $P'$) and $(a, b)$ present in $\mathrm{rel}^P(c_{xy})$ is removed from $\mathrm{rel}^P(c_{xy})$. When enforcing sDC (with $\mathsf{learn}^{full}$) and when no binary constraint exists between $x$ and $y$, a new constraint is created. This constraint enforces the set of nogoods corresponding to DC-inconsistent pairs of values involving $(x, a)$ and variable $y$. The new constraint accepts every pair of values except those that have been identified as conflicts. Notice that we know that there is at least one conflict since $\mathrm{dom}^P(y) \neq \mathrm{dom}^{P'}(y)$ at line 8 of Algorithm 2.

**Proposition 30.** *Algorithm sCDC1 enforces sCDC on binary CNs and SAC+CDC on non-binary CNs; Algorithm sDC1 enforces sDC.*

*Proof.* First, any inference performed by Algorithm 2 at Line 5 or by $\mathsf{learn}^{part}/\mathsf{learn}^{full}$ is correct: such inferences correspond to clearly identified SAC-inconsistent values and (C)DC-inconsistent pairs of values. Any inference directly performed by Algorithm 1 at Lines 1 and 7 is also safe because it corresponds to removing GAC-inconsistent values; remember that the overall algorithm enforces *strong* (C)DC, or SAC+CDC (and SAC is stronger than GAC). The fact that all possible inferences are performed is guaranteed by the fact that each time a revision is effective, the marker used in Algorithm 1 is updated; see Line 9. Also, any inference performed with respect to a pair $(x, a)$ has no effect on $GAC(P|_{x=b})$, where $b$ is any other value in the domain of the variable $x$. Indeed, when $b$ is assigned to $x$, all other values in the current domain of $x$ are automatically removed. Combined with the enforcement of GAC, the assignment of a new value to $x$ makes previous inferences related to other values of $x$ without any effect. This is the reason why we can iterate over all values of $x$ at Line 2 of Algorithm 2 in a unique pass. □

One *pass* of Algorithm 1 means calling reviseVariable exactly once per variable.

**Proposition 31.** *One pass of Algorithm 1 has a worst-case time complexity in $O(enrd^{r+1})$ with $\mathsf{learn}^{part}$ (i.e., for sCDC1) and in $O(enrd^{r+1} + n^3d^3)$ with $\mathsf{learn}^{full}$ (i.e., for sDC1).*

*Proof.* The optimal worst-case time complexity of enforcing GAC is $O(erd^r)$ (Mohr & Masini, 1988). The worst-case time complexity of Lines 8–10 of Algorithm 2 is $O(nd)$ with both learn methods. Besides,

- with $\mathsf{learn}^{part}$, no new constraint is inserted in $P$, the worst-case time complexity of one pass of Algorithm 1 is $O(nd \cdot (erd^r + nd))$. This reduces to $O(enrd^{r+1})$ with the very weak assumption that $n \leq er$ (otherwise, some variables would not be involved in any constraint);

- with learn$^{full}$, we have to consider that $O(n^2)$ additional binary constraints may be added by the algorithm. So, we obtain $O(enrd^{r+1} + n^3d^3)$. □

For binary constraints ($r = 2$, which entails $e < n^2$), we obtain:

**Corollary 1.** *For binary CNs, one pass of Algorithm 1 admits a worst-case time complexity in $O(end^3)$ with* learn$^{part}$ *and in $O(n^3d^3)$ with* learn$^{full}$.

When $P$ is not already s(C)DC-consistent, several passes of Algorithm 1 are necessary. Thus,

- with learn$^{part}$, the number of passes is bounded by $O(ed^2)$; one tuple being removed at each new pass. The worst-case time complexity of Algorithm 1 is then in $O(e^2nrd^{r+3})$ ($O(e^2nd^5)$ for binary CNs);

- with learn$^{full}$, the number of passes is bounded by $O(n^2d^2)$; one tuple being removed at each new pass. The resulting worst-case time complexity is then $O(en^3rd^{r+3} + n^5d^5)$ ($O(n^5d^5)$ for binary CNs).

The overall time complexity of Algorithm 1 seems to be rather high but we have observed that very often a fixed point is quickly reached in practice (i.e., the number of passes empirically tends to be constant). We can also note that for sDC1 (i.e., Algorithm 1 with learn$^{full}$), it is possible to limit the cost of enforcing GAC at each singleton check. Indeed, when a singleton check on a pair $(x, a)$ is performed, every nogood of size 2 including $(x, a)$ is identified and recorded in the CN $P$. This means that just after the last instruction of each iteration of the *foreach* loop starting at Line 2 of Algorithm 2, $P$ is such that after assigning again $a$ to $x$, applying forward checking (any non-binary version, Bessiere, Meseguer, Freuder, & Larrosa, 2002) is enough to enforce GAC: we only need to consider binary constraints involving $x$ and delete values that are not consistent with $(x, a)$. This is studied in a paper by Lecoutre et al. (2007b).

Finally, it is worthwhile to mention that Algorithm 1 does not require specific data structure. The only data structures required are those of the underlying (G)AC algorithm(s) and those for the representation of CNs. If $e_b$ denotes the number of binary constraints in a given CN, then sCDC1 may require $O(e_bd^2)$ additional space to store new nogoods (if binary constraints were initially given in intension for example). sDC1 require $O(n^2d^2)$ additional space, which may be a serious drawback for solving certain problems.

## 6. Experimental Results

To show the practical interest of strong second-order consistencies, and in particular s(C)DC, we conducted several extensive experiments on Oracle Java 6 VMs running on a cluster of Intel Xeon 3.0 GHz with 1 GiB of RAM under Linux. To do this, we implemented sCPC, sCDC and sDC algorithms in our constraint solver AbsCon. These implementations are sCPC8, directly derived from PC8 (Chmeiss & Jégou, 1998), sCDC1 and sDC1 that correspond to Algorithm 1 with learn$^{part}$ and learn$^{full}$, respectively. Some refinements of the algorithm sDC1, as proposed by Lecoutre et al. (2007b), are also possible but will not be considered in this paper, mainly because such optimizations are rather marginal (a speed-up of 10 % on some problems) with respect to our main concern: showing that enforcing s(C)DC before search may pay off.

The AbsCon solver also features SAC preprocessing algorithms. We used SAC1 (Bessiere & Debruyne, 2005) and SAC3 (Lecoutre & Cardon, 2005) algorithms in some experiments. In AbsCon, the algorithms used to enforce (G)AC are $AC3^{bit+rm}$ (Lecoutre & Vion, 2008) for binary constraints, $GAC3^{rm}$ (Lecoutre & Hemery, 2007) for non-binary constraints defined in intension, and STR2 (Lecoutre, 2008) for non-binary constraints in extension. Besides, when possible, an optimization based on reasoning from the cardinality of conflict sets (Boussemart, Hemery, Lecoutre, & Sais, 2004b) was used. Of course, SAC1, SAC3, sCDC1 and sDC1 also benefit from the efficiency of these underlying (G)AC algorithms.

## 6.1 Preprocessing Performance on Random Problems

We first evaluate the performance of various consistencies and algorithms on random instances. Random CNs are generated using Model B (Gent, MacIntyre, Prosser, Smith, & Walsh, 2001) to comply to five parameters: the number of variables $n$, the size of the domains $d$, the arity of the constraints $r$ ($r = 2$ in our experiment), the density $\delta$,[7] and the tightness $t$ (proportion of tuples forbidden by each constraint).

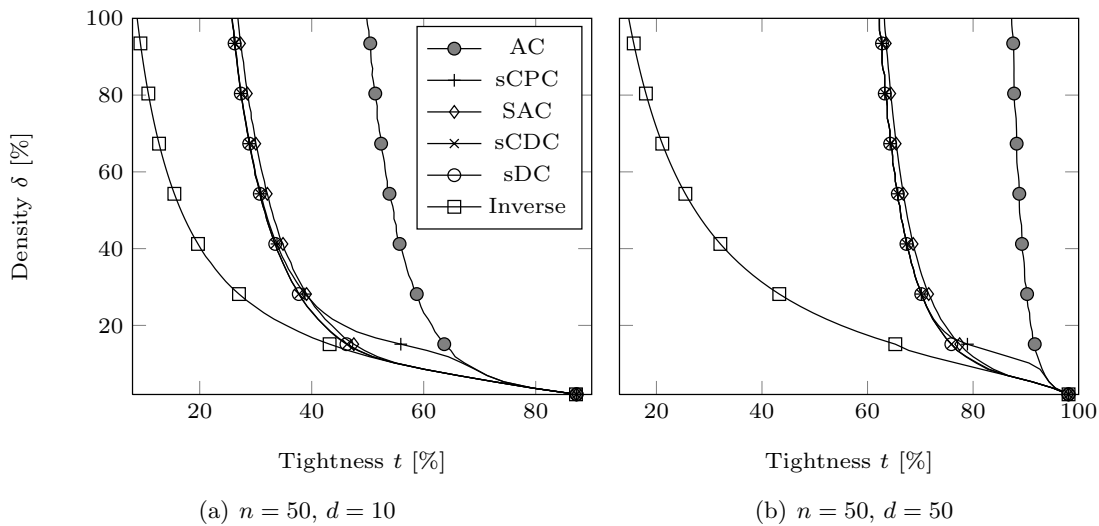

(a) $n = 50$, $d = 10$

(b) $n = 50$, $d = 50$

Figure 14: Phase transition of various consistencies for random binary CNs.

Figure 14 gives some quantitative information about the relative strength of AC, sCPC, SAC, sCDC and sDC (= sPC) enforced on random binary CNs with $n = 50$ variables and $d \in \{10, 50\}$ values per domain. Each plot in the figures represents the position of the phase transition of a given consistency for the generated instances; there are 50 instances generated for each $(\delta, t)$ pair, and we recall that the phase transition of a consistency $\phi$ occurs when 50 % of the generated instances are detected as unsatisfiable when enforcing $\phi$. Phase transition for inverse consistency is also plotted as a baseline: a CN is inverse-consistent, or $(1, n)$-consistent, iff each value belongs to at least one solution. The obtained results show that sCPC is rather close to sCDC on such random instances, except when the tightness $t$ ranges between 40% and 70%. Surprisingly, the difference between sCDC and sDC is so weak that it is not even visible on the graphs. Also, we can see that second-order consistencies become closer to inverse consistency when the size of the domains is lower.

---

7. The density determines the number of constraints $e = \delta \cdot \binom{n}{r}$

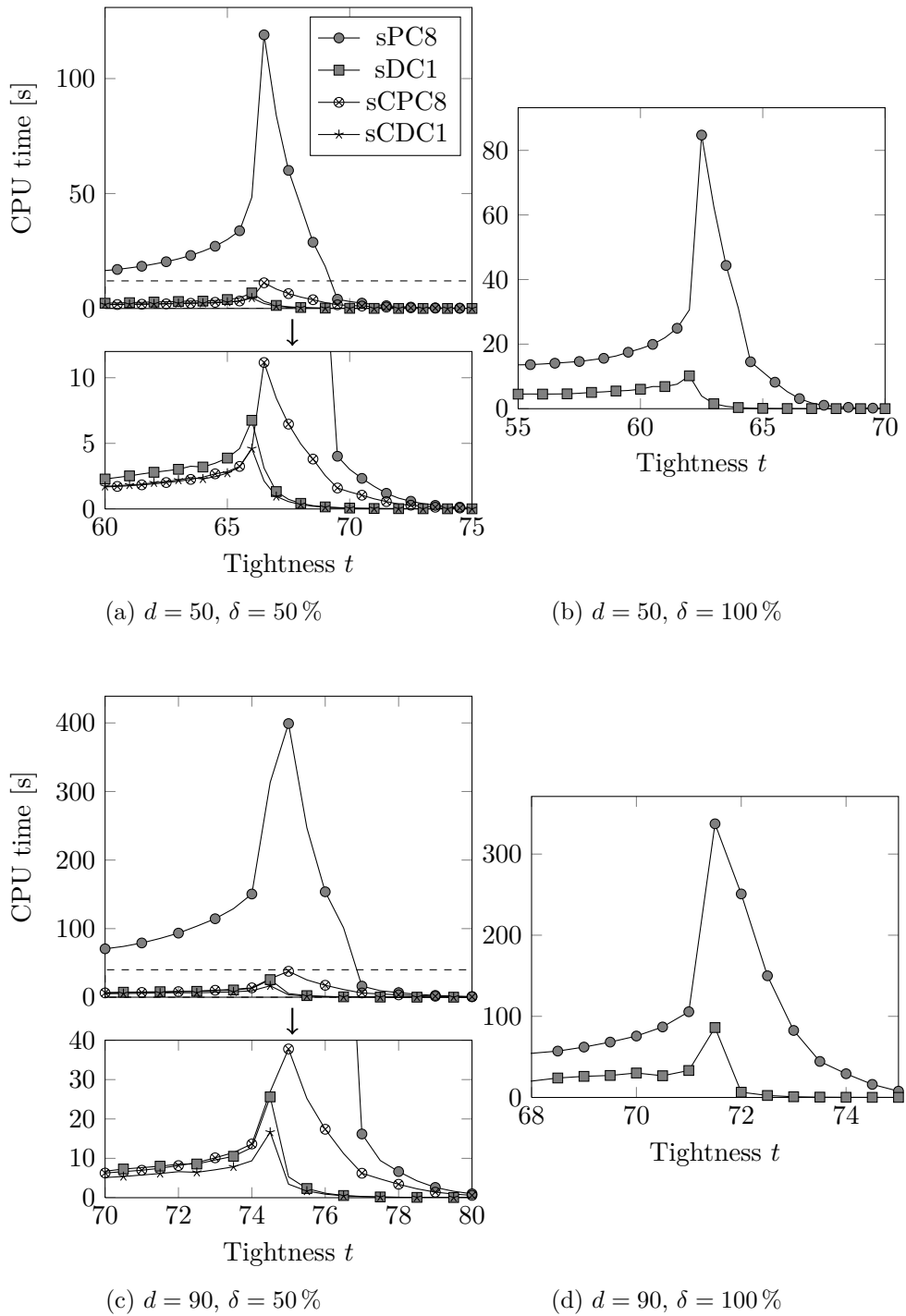

Figure 15: Mean CPU time (in seconds) for enforcing various second-order consistencies on binary random CNs ($n = 50$).

Figure 15 compares the CPU times required to establish sPC, sDC, sCPC and sCDC, on binary random CNs, using the algorithms sPC8, sDC1, sCPC8 and sCDC1, respectively. For each value of $t$, 50 instances have been generated, with either 50 values (topmost figures) or 90 values (bottommost figures) per variable. Leftmost pictures represent performances on CNs with density $\delta = 50\%$ and rightmost pictures on CNs with complete constraint graphs. In the latter case, all consistencies are equivalent; this is why only plots for sDC1 and sPC8 are given. The results are quite informative. On the one hand, they illustrate the very high performance of s(C)DC algorithms with respect to state-of-the-art s(C)PC algorithms, with a performance gap often over one order of magnitude on dense problems. On the other hand, the obtained results show that establishing sDC is not much harder than enforcing sCDC.

## 6.2 Impact of Second-Order Consistency Preprocessing on MAC

Now, we turn to complete search algorithms. One of the most popular (systematic) search algorithms to solve CNs is called MAC (Sabin & Freuder, 1994). MAC interleaves inference and search since at each step of a depth-first exploration with backtracking, (generalized) arc consistency is maintained.[8] At each step of search, MAC uses a variable ordering heuristic to select the next variable to be instantiated. Two representative variable ordering heuristics are the dynamic heuristic *dom/ddeg* (Bessière & Régin, 1996) and the adaptive heuristic *dom/wdeg* (Boussemart, Hemery, Lecoutre, & Sais, 2004a). Today, enforcing (strong) second-order consistencies during search (basically, after each variable assignment) seems to be unrealistic. However, enforcing such consistencies in a preprocessing stage and before running MAC is an issue that deserves to be addressed; in this section, we attempt to provide some insight in this regard. If $\Phi$ denotes the consistency enforcing algorithm applied at preprocessing, this is denoted by $\Phi$-MAC. The experimentation that we have conducted consists in comparing MAC, SAC1-MAC (and also SAC3-MAC), sCPC8-MAC, sCDC1-MAC and sDC1-MAC; from now on called *variants* of MAC. It allows us to assess the practical impact of enforcing (strong) second-order consistencies at preprocessing on many series of structured (i.e., not random) binary and non-binary instances. These series as well as their description can be found at `http://www.cril.fr/~lecoutre/benchmarks.html`. We briefly introduce the main features of these series (listed in alphabetical order).

**aim:** series of instances with Boolean variables and ternary constraints. Initially, they were generated as Boolean formulas in DIMACS CNF (Conjunctive Normal Form) format.

**bqwh:** series of satisfiable balanced quasi-group instances with holes (Gomes & Shmoys, 2002); domains of variables are small (usually, a few values) and constraints are binary.

---

8. For simplicity, we shall use the acronym MAC whatever the arity of the constraints is.

**composed:** series of instances randomly composed of a main (under-constrained) fragment and some auxiliary fragments (Lecoutre, Boussemart, & Hemery, 2004); there are 10 values per domain and constraints are binary.

**driver:** series of planning instances with costs converted from WCSP (Weighted CSP) to CSP; domains of variables are small (usually, a few values) and constraints are binary.

**ehi:** series of SAT instances converted into CSP using the dual method as described by Bacchus (2000); there are 7 values per domain and constraints are binary.

**langford:** series from the (generalized version of the) Langford problem; domains of variables are usually not small (up to 150 values) and constraints are binary.

**os-taillard:** series of Open-Shop scheduling instances generated by J. Vion from Taillard's paper (1993); domains are large and constraints are binary.

**primes:** series of instances involving prime numbers generated by M. van Dongen; domains of variables are not small and some constraints are non-binary.

**qcp / qwh:** series from the Quasi-group Completion Problem (QCP) and the Quasi-group With Holes problem (QWH); domains are not very large and constraints are binary.

**queensKnights:** series of academic instances with queens and knights to be put on a chessboard (Boussemart et al., 2004a); some domains are large and constraints are binary.

**radar:** realistic radar surveillance problems generated following the model of the Swedish institute of computer science (SICS); domains are small and constraints non-binary.

**renault-mod:** series of instances generated by K. Stergiou from a Renault Megane configuration problem; domains are diverse and constraints are non-binary.

**sadeh:** series containing the five sets of job-shop scheduling instances studied by Sadeh & Fox (1996); domains are large and constraints are binary.

**scens11:** series containing hard variants of the original scen11 instance from the Radio Link Frequency Assignment Problem (RLFAP, Cabon et al., 1999); domains are not small and constraints are binary.

**series:** series from the all-interval series problem; domains are diverse and constraints are binary or ternary.

**ruler:** series from the Golomb Ruler problem; domains are not small and constraints are binary, ternary or quaternary.

**tsp:** series generated by R. Szymanek from the Travelling Salesperson problem; domains are large and constraints are binary or ternary.

Tables 1 and 3 show the results that we have obtained with the different variants of MAC (the third column is for MAC alone) equipped with the variable ordering heuristics $dom/ddeg$ and $dom/wdeg$, respectively. For each series, the number $nb_s$ of instances solved by the five variants of MAC within the alloted time (20 minutes) as well as the total number $nb$ of instances are given under the form $nb_s/nb$ in column #Inst. The mean CPU times displayed in these two tables is computed from the $nb_s$ instances identified for each series; when a variant solves some additional instances, this is indicated between brackets preceded by the $+$ symbol. It is interesting to note that the SAC3 algorithm (Lecoutre & Cardon, 2005) sometimes permits to discover "lucky" solutions during preprocessing due to its greedy nature. When this happens, we show the increased number of solved instances behind /. For example, in Table 1, the first row indicates that 12 instances out of a set

| Series | #Inst | MAC with at preprocessing $\Phi$ = | | | | |
|---|---|---|---|---|---|---|
| | | – | SAC1/3 | sCPC8 | sCDC1 | sDC1 |
| aim-100 | 12/24 | 100 | (+2) 65.6 | 106 | (+2) 68.1 | (+3) **54.2** |
| aim-200 | 6/24 | 262 | (+7) 263 | 273 | (+7) 265 | (+10) **16.4** |
| bqwh-15 | 100/100 | 1.77 | 1.63 | 1.37 | **1.19** | 3.5 |
| bqwh-18 | 99/100 | 55.9 | 54.6 | (+1) 33.3 | (+1) 29.3 | (+1) **26.3** |
| composed-25 | 15/50 | 121 | (+35) **0.89** | (+35) 0.97 | (+35) 1.1 | (+35) 1.2 |
| composed-75 | 3/40 | 0.6 | (+37) **0.73** | (+37) 0.8 | (+37) 0.8 | (+37) 1.1 |
| driver | 7/7 | 44.0 | 11.4 | 9.74 | **8.48** | 53.9 |
| ehi-85 | 99/100 | 197 | (+1) 1.47 | (+1) 1.84 | (+1) **1.32** | (+1) 1.41 |
| ehi-90 | 94/100 | 251 | (+6) 1.46 | (+6) 1.88 | (+6) **1.3** | (+6) 1.4 |
| langford-2 | 16/24 | **1.44** | 1.73 | 36.9 | 3.77 | 3.29 |
| langford-3 | 16/24 | 49.6 | 47.4 | 38.9 | **25.7** | 26.0 |
| langford-4 | 14/24 | (+1) 25.0 | 29.0 | (+1) 38.0 | (+1) 21.8 | (+1) **21.7** |
| os-taillard-4 | 29/30 | (+1) 13.6 | (+1) **1.85** | (+1) 5.79 | (+1) 2.15 | 3.10 |
| os-taillard-5 | 10/30 | (+5) 13.1 | (+9/10) 6.58 | (+9) 103 | (+8) 65.2 | (+10) **94.8** |
| os-taillard-7 | 6/30 | (+2) 64.5 | (+2/3) 75.2 | 247 | 77.0 | (+3) **118** |
| primes-10 | 25/32 | 45.3 | 92.4 | 45.0 | 91.1 | (+1) **86.2** |
| primes-15 | 20/32 | (+2) 3.01 | 12.6 | (+2) **3.09** | 15.8 | (+2) 13.9 |
| primes-20 | 20/32 | 3.60 | (+2) 76.6 | 3.77 | (+2) 87.9 | (+2) **73.4** |
| qcp-10 | 14/15 | (+1) **18.1** | (+1) 18.6 | (+1) 19.8 | (+1) 18.4 | 18.4 |
| qcp-15 | 5/15 | (+4) **212** | (+4) 214 | (+3) 172 | (+3) 168 | (+1) 129 |
| qcp-20 | 0/15 | – | – | – | – | – |
| queensKnights | 6/18 | 97.9 | (+11) 0.74 | (+6) 0.82 | (+11) **0.64** | (+11) 0.69 |
| qwh-10 | 10/10 | 0.77 | **0.76** | 0.93 | 0.82 | 1.10 |
| qwh-15 | 10/10 | 5.95 | 11.7 | 8.56 | 7.2 | **3.34** |
| qwh-20 | 0/15 | – | – | (+3) – | (+4) – | (+8) – |
| radar-8-24 | 34/50 | (+4) 12.9 | (+5) 20.9 | (+4) 13.7 | (+5) **14.2** | (+4) 3.09 |
| radar-8-30 | 35/50 | (+1) 3.5 | (+4) 35.1 | (+1) 3.75 | (+3) 3.6 | (+4) **20.2** |
| radar-9-28 | 12/50 | 90.7 | (+5) 1.58 | 93.5 | (+5) 1.35 | (+5) **1.33** |
| renault-mod | 46/50 | (+1) 36.4 | (+2) **4.05** | (+1) 36.8 | (+2) 9.30 | (+1) 20.6 |
| sadeh | 16/46 | (+11) 4.52 | (+11/**22**) 5.71 | (+11) 10.1 | (+12) 4.10 | (+**13**) **78.8** |
| scens11 | 0/11 | – | – | (+3) – | (+3) – | (+3) – |
| series | 10/25 | 44.5 | 45.9 | 47.6 | 43.6 | (+**15**) **0.83** |
| ruler | 17/28 | (+1) 77.3 | (+1) 70.8 | (+1) 70.1 | 88.9 | (+1) **62.4** |
| tsp-20 | 15/15 | **6.63** | 8.27 | 8.08 | 19.6 | 63.7 |
| tsp-25 | 13/15 | (+2) 84.8 | (+2) 72.2 | (+2) 97.1 | (+2) **63.9** | 223 |
| | | +36 | +148/161 | +129 | +152 | +178 |

Table 1:  Mean cpu time (in seconds) to solve instances of different series (time-out of 1,200 s per instance) with $\Phi$-MAC-*dom/ddeg*.

| Instances | | MAC with at preprocessing Φ = | | | |
|---|---|---|---|---|---|
| | | − | sCPC8 | sCDC1 | sDC1 |
| **aim-100-1-6-sat** | cpu (pcpu) | 50.9 (0) | 46.6 (0.01) | **0.38 (0.03)** | 0.47 (0.12) |
| | mem | 25 M | 25 M | 25 M | 25 M |
| | nodes | 655 K | 655 K | 100 | 100 |
| | del v - del t | 0 - 0 | 0 - 0 | 100 - 0 | 100 - 231 (200) |
| **bqwh-18-141-30** | cpu (pcpu) | 180 (0.01) | 66.2 (0.11) | 92.2 (0.20) | **49.5 (0.31)** |
| | mem | 31 M | 31 M | 31 M | 35 M |
| | nodes | 1,682 K | 643 K | 845 K | 443 K |
| | del v - del t | 6 - 0 | 6 - 182 | 9 - 207 | 9 - 585 (296) |
| **driver-02c-sat** | cpu (pcpu) | **2.34 (0.03)** | 3.68 (1.23) | 3.66 (2.33) | 15.0 (13.3) |
| | mem | 35 M | 43 M | 39 M | 66 M |
| | nodes | 5,278 | 3,277 | 988 | 505 |
| | del v - del t | 64 - 0 | 64 - 9,439 | 343 - 5,336 | 343 - 31 K (7 K) |
| **driver-09-sat** | cpu (pcpu) | 222 (0.05) | 52.7 (20.9) | **10.3 (8.2)** | 61.1 (58.6) |
| | mem | 57 M | 135 M | 73 M | 217 M |
| | nodes | 215 K | 14,284 | 650 | 650 |
| | del v - del t | 162 - 0 | 162 - 63,163 | 2,175 - 22,590 | 2 K - 117 K (19 K) |
| **langford-4-15** | cpu (pcpu) | 254 (0.03) | 287 (9.29) | **215 (7.61)** | **215 (7.45)** |
| | mem | 32 M | 52 M | 40 M | 40 M |
| | nodes | 1,056 K | 197 K | 197 K | 197 K |
| | del v - del t | 1,620 - 0 | 1,620 - 517 K | 1,620 - 517 K | 1,620 - 517 K (0) |
| **os-taillard-5-95-2** | cpu (pcpu) | > 1,200 | > 1,200 | **9.06 (6.56)** | 15.1 (12.1) |
| | mem | | | 32 M | 40 M |
| | nodes | | | 3,204 | 2,562 |
| | del v - del t | | | 570 - 384 K | 570 - 1 M (175) |
| **primes-15-20-2-3** | cpu (pcpu) | 1.81 (0.32) | **1.77 (0.27)** | 16.5 (15.4) | 22.1 (21.4) |
| | mem | 29 M | 29 M | 29 M | 29 M |
| | nodes | 122 | 122 | 104 | 101 |
| | del v - del t | 657 - 0 | 657 - 0 | 766 - 0 | 766 - 2,491 (46) |
| **primes-20-20-3-3** | cpu (pcpu) | > 1,200 | > 1,200 | 188 (178) | **153 (141)** |
| | mem | | | 29 M | 29 M |
| | nodes | | | 175 | 181 |
| | del v - del t | | | 602 - 0 | 653 - 18,804 (96) |
| **scen11-f12** | cpu (pcpu) | >1200 | 15.3 (11.5) | **7.95 (4.84)** | 13.7 (10.4) |
| | mem | | 53 M | 49 M | 96 M |
| | nodes | | 240 | 18 | 18 |
| | del v - del t | | 6,324 - 419 K | 6,768 - 306 K | 7 K - 351 K (3 K) |
| **renault-mod-8** | cpu (pcpu) | 24.0 (0.05) | 24.5 (0.04) | 8.05 (6.07) | **7.84 (6.8)** |
| | mem | 67 M | 67 M | 67 M | 101 M |
| | nodes | 179 K | 179 K | 0 | 0 |
| | del v - del t | 22 - 0 | 22 - 0 | 95 - 0 | 72 - 11 K (1,719) |
| **series-12** | cpu (pcpu) | 12.2 (0.01) | 12.9 (0.10) | 12.0 (0.26) | **0.81 (0.33)** |
| | mem | 29 M | 29 M | 29 M | 29 M |
| | nodes | 106 K | 106 K | 106 K | 23 |
| | del v - del t | 0 - 0 | 0 - 0 | 0 - 0 | 0 - 310 (110) |

Table 2: Detailed results on various instances (time-out of 1,200 s per instance) with Φ-MAC-*dom/ddeg*.

| Series | #Inst | MAC with at preprocessing $\Phi$ = | | | | |
| --- | --- | --- | --- | --- | --- | --- |
| | | − | SAC1/3 | sCPC8 | sCDC1 | sDC1 |
| aim-100 | 24/24 | 0.97 | 0.90 | 1.05 | 1.01 | **0.84** |
| aim-200 | 24/24 | 6.86 | 1.94 | 7.13 | 3.33 | **1.25** |
| bqwh-15 | 100/100 | 0.99 | 1.12 | 1.05 | **0.98** | 1.39 |
| bqwh-18 | 100/100 | 5.37 | 3.87 | 4.04 | 3.93 | **3.71** |
| composed-25 | 50/50 | 0.70 | **0.69** | 0.82 | 0.84 | 0.82 |
| composed-75 | 40/40 | 1.02 | **0.86** | 0.94 | 1.02 | 0.96 |
| driver | 7/7 | **3.14** | 10.7 | 8.43 | 8.2 | 55.7 |
| ehi-85 | 100/100 | 1.96 | 1.44 | 1.87 | **1.30** | 1.31 |
| ehi-90 | 100/100 | 1.97 | 1.57 | 2.00 | 1.40 | **1.36** |
| langford-2 | 16/24 | **1.47** | 1.75 | 39.8 | 3.36 | 3.32 |
| langford-3 | 16/24 | 56.2 | 57.6 | 46.4 | 34.6 | **32.9** |
| langford-4 | 14/24 | **26.5** | (+0/1) 29.1 | 44.0 | 27.5 | 28.3 |
| os-taillard-4 | 30/30 | **1.08** | 1.74 | 5.66 | 1.89 | 2.57 |
| os-taillard-5 | 28/30 | 40.5 | (+1) 65.0 | 52.0 | (+2) **45.5** | (+2) 48.4 |
| os-taillard-7 | 11/30 | (+3) 70.0 | (+3/9) 76.5 | (+1) 266 | (+1) 103 | (+4) **145** |
| primes-10 | 26/32 | 71.1 | (+0/2) 73.6 | **71.9** | 78.5 | 97.6 |
| primes-15 | 22/32 | (+1) 2.43 | (+2/1) 27.2 | (+1) 2.34 | (+2) 16.2 | (+2) **14.5** |
| primes-20 | 21/32 | (+2) 13.4 | (+1/2) 82.9 | (+2) **13.8** | (+1) 91.0 | (+2) 75.9 |
| qcp-10 | 15/15 | **0.71** | 0.84 | 0.88 | 0.77 | 0.91 |
| qcp-15 | 15/15 | 43.1 | **16.8** | 62.8 | 60.8 | 31.1 |
| qcp-20 | 0/15 | (+3) − | (+3/4) − | (+3) − | (+4) − | (+3) − |
| queensKnight | 7/18 | (+1) 2.57 | (+10) 0.83 | (+5) 0.88 | (+10) 0.64 | (+10) **0.66** |
| qwh-10 | 10/10 | **0.69** | 0.81 | 1.41 | 0.76 | 1.01 |
| qwh-15 | 10/10 | 1.67 | 2.27 | 2.54 | 1.72 | **1.64** |
| qwh-20 | 10/10 | 152 | 125 | 96.0 | 73.2 | **41.4** |
| radar-8-24 | 50/50 | 26.6 | 27.6 | 26.5 | 27.9 | **26.4** |
| radar-8-30 | 50/50 | **1.43** | 37.3 | 1.55 | 2.11 | 2.10 |
| radar-9-28 | 27/50 | (+1) 90.2 | (+2/10) **37.9** | (+1) 88.9 | (+2) 71.5 | (+2) 67.5 |
| renault-mod | 50/50 | **1.65** | 2.72 | 1.88 | 7.94 | 11.8 |
| sadeh | 37/46 | (+1) 13.1 | (+1/7) 25.1 | (+1) 18.8 | (+3) 33.2 | (+6) **48.8** |
| scens11 | 9/12 | 95.4 | (+0/1) **54.6** | 109 | 81.6 | 172 |
| series | 10/25 | (+1) 12.6 | (+0/1) 1.76 | (+1) 13.5 | 16.3 | (+15) **0.79** |
| ruler | 19/28 | (+2) **16.2** | (+1/0) 46.3 | (+2) 24.0 | (+1) 66.0 | (+1) 41.6 |
| tsp-20 | 15/15 | **3.91** | 6.09 | 4.48 | 18.1 | 69.1 |
| tsp-25 | 13/15 | (+2) **20.9** | (+2) 28.4 | (+2) 25.4 | (+2) 59.0 | 271 |
| | | +17 | +26/51 | +19 | +28 | +47 |

Table 3: Mean cpu time (in seconds) to solve instances of different series (time-out of 1,200 s per instance) with $\Phi$-MAC-*dom/wdeg*.

| Instances | | MAC with preprocessing Φ = | | | |
|---|---|---|---|---|---|
| | | – | sCPC8 | sCDC1 | sDC1 |
| e0ddr1-4 | cpu (pcpu) | 1.35 (0.01) | 5.25 (4.0) | 3.05 (1.8) | 8.57 (7.3) |
| | mem | 32 M | 32 M | 32 M | 60 M |
| | nodes | 50 | 50 | 50 | 50 |
| | del v - del t | 0 - 0 | 0 - 0 | 0 - 0 | 0 - 531 K (927) |
| e0ddr1-5 | cpu (pcpu) | 125 (0.01) | 121 (4.17) | 3.19 (1.94) | 9.08 (7.7) |
| | mem | 32 M | 32 M | 32 M | 60 M |
| | nodes | 1,245 K | 1,245 K | 75 | 56 |
| | del v - del t | 0 - 0 | 0 - 0 | 0 - 7,006 | 0 - 529 K (922) |
| qcp-15-120-2 | cpu (pcpu) | 93 (0.04) | 34.8 (0.36) | 36.0 (0.33) | 2.41 (0.45) |
| | mem | 32 M | 36 M | 36 M | 40 M |
| | nodes | 955 K | 308 K | 333 K | 706 |
| | del v - del t | 1,276 - 0 | 1,276 - 302 | 1,282 - 177 | 1,282 - 418 (189) |
| qcp-20-187-11 | cpu (pcpu) | 2.33 (0.15) | > 1,200 | 2.4 (0.58) | 2.66 (0.98) |
| | mem | 37 M | | 45 M | 81 M |
| | nodes | 4,898 | | 3,435 | 2,099 |
| | del v - del t | 2,913 - 0 | | 2,930 - 325 | 2,930 - 1,091 (583) |
| scen11-f8 | cpu (pcpu) | 6.93 (0.06) | 31.2 (26.5) | 8.39 (4.6) | 12.3 (6.85) |
| | mem | 37 M | 57 M | 53 M | 89 M |
| | nodes | 15,640 | 2,214 | 2,214 | 6,106 |
| | del v - del t | 4,992 - 0 | 4,992 - 365 K | 4,992 - 366 K | 4,992 - 389 K (757) |
| scen11-f6 | cpu (pcpu) | 37.2 (0.05) | 74.8 (33.9) | 45.9 (4.47) | 74.4 (6.46) |
| | mem | 37 M | 57 M | 53 M | 89 M |
| | nodes | 204 K | 167 K | 167 K | 260 K |
| | del v - del t | 3,660 - 0 | 3,660 - 415 K | 3,660 - 416 K | 3,660 - 443 K (757) |
| series-14 | cpu (pcpu) | 93.2 (0.01) | 101 (0.16) | 125 (0.2) | 0.91 (0.41) |
| | mem | 29 M | 29 M | 29 M | 29 M |
| | nodes | 689 K | 689 K | 914 K | 27 |
| | del v - del t | 0 - 0 | 0 - 0 | 0 - 0 | 0 - 444 (156) |
| series-25 | cpu (pcpu) | > 1,200 | > 1,200 | > 1,200 | 3.16 (2.34) |
| | mem | | | | 41 M |
| | nodes | | | | 49 |
| | del v - del t | | | | 0 - 1,610 (552) |
| tsp-20-75 | cpu (pcpu) | 3.32 (0.13) | 4.04 (0.36) | 26.9 (25.7) | 90.0 (81.3) |
| | mem | 35 M | 51 M | 47 M | 197 M |
| | nodes | 6,926 | 6,926 | 1,631 | 2,064 |
| | del v - del t | 21 K - 0 | 21 K - 0 | 22 K - 27 K | 22 K - 895 K (1.4 K) |
| os-taill-7-100-8 | cpu (pcpu) | 165 (0.01) | 378 (361) | 24.8 (8.4) | 31.2 (8.8) |
| | mem | 49 M | 53 M | 49 M | 53 M |
| | nodes | 667 K | 42,499 | 42,499 | 51,403 |
| | del v - del t | 0 - 0 | 0 - 58 | 0 - 58 | 0 - 60 (1) |
| os-taill-7-105-9 | cpu (pcpu) | 9.43 (0.01) | 336 (256) | 86.2 (8.4) | 16.2 (9.4) |
| | mem | 46 M | 50 M | 50 M | 50 M |
| | nodes | 4,210 | 179 K | 179 K | 49 |
| | del v - del t | 0 - 0 | 0 - 4,328 | 0 - 4,328 | 0 - 11,072 (20) |

Table 4: Detailed results on various instances (time-out of 1,200 s per instance) with Φ-MAC-*dom/wdeg*.

of 24 instances from series aim-200 have been solved by each of the five variants of MAC. sDC1-MAC solves these 12 instances with a mean CPU time computed at 54.2 seconds, and additionally solves 3 other instances of the series (within 20 minutes). Note that for each series, the mean CPU time of the best variant of MAC is printed in bold face: the best variant is the one that solves the highest number of instances within 20 minutes and in case of equality, the one that has the smallest mean CPU time.

Tables 2 and 4 provide details on solving various instances when enforcing second-order consistencies at preprocessing. For each problem instance, the total CPU time to solve it is given (this is > 1,200 when the instance cannot be solved within 20 minutes) as well as the time taken to enforce the consistency at preprocessing (*pcpu* between brackets). Additional information concern the memory requirement (expressed in MiB), the number of explored nodes and the number of values and tuples deleted at preprocessing (*del v – del t*). For sDC1, we also put the number of new binary constraints between brackets. For example, in Table 2, for solving the instance aim-100-1-6-sat, sDC1-MAC needs 0.47 s (0.12 s at preprocessing), 25 MiB of memory, explores 100 nodes, deletes 100 values and 231 tuples at preprocessing (while adding 200 new binary constraints).

Table 1 clearly shows that when the heuristic *dom/ddeg* is used, there is a real interest of making a strong propagation effort during preprocessing on many problems. Although MAC remains the most efficient approach on a few series (e.g., langford-2 and tsp-20), sCDC1-MAC and sDC1-MAC are proved here to be more robust than MAC. For example, on series aim-200, radar-9-28 and series, sDC1-MAC largely outperforms MAC. Overall, sDC1-MAC solves $178 - 36 = 142$ more instances than MAC. sCDC1-MAC and SAC1/3-MAC are comparable in terms of the number of solved instances. However, on some series (e.g., langford-3 and bqwh-18), sCDC1-MAC is clearly better than SAC1/3-MAC, even if the reverse is true for some other series (e.g., renault-mod and sadeh). It is also interesting to note that sCPC8-MAC is almost always outperformed by sCDC1-MAC and sDC1-MAC. This is an expected result, as sCPC is weaker than sCDC (and *a fortiori* sDC), and s(C)DC algorithms benefit of underlying highly optimized (G)AC algorithms. On some series, adding new binary constraints in order to collect all identified nogoods of size 2 may be counterproductive. The conservative consistency sCDC is then a better option than sDC. For example, this is the case for driver and tsp-25 series.

The results given in Table 2 show the dramatic effect of a strong consistency at preprocessing on some instances. For example, after enforcing sCDC or sDC, MAC is able to directly find a solution to aim-100-1-6-sat. This is also the case for sDC1-MAC on series-12. However, sDC1 may require a significant amount of additional memory (to record new constraints). This may explain the relative inefficiency of sDC1-MAC on instances driver-02c-sat, driver-09-sat and scen11-f12, compared to sCDC1-MAC. Note that for renault-mod-8, sCDC1 and sDC1 alone are sufficient to detect unsatisfiability (the number of visited nodes is 0). However, the unsatisfiability of this instance is proved differently (i.e., following different propagation paths), which can explain why 95 and 72 values are respectively deleted when sCDC and sDC are enforced.

Table 3 confirms that MAC equipped with the heuristic *dom/wdeg* is far more robust than *dom/ddeg*, as initially claimed by Boussemart et al. (2004a). As a result, on some series, enforcing a strong consistency (i.e. a consistency stronger than GAC) at preprocessing has a more limited impact. Nevertheless, overall, sDC1-MAC still increases the robustness of the solver. Concerning SAC3-MAC, it is important to note that its good behaviour is due to its opportunistic mechanism of finding solutions, and not to its inference capability as shown by the results of SAC1-MAC: SAC1-MAC solves 26 additional instances against 51 for SAC3-MAC. If sCDC1-MAC is outperformed by sDC1-MAC, note that on some series, it remains a good option because it does not perturbate the heuristic by adding new constraints (e.g., compare the number of nodes for solving instances `scen11-f6` and `scen11-f8` in Table 4) and its overhead at preprocessing is limited (compare for example the preprocessing time of sCDC1 and sDC1 for `e0ddr1-4` and `tsp-20-75` in Table 4).

Which lessons can be learned from this experimental study? A first one is that enforcing s(C)DC is most of the time far more efficient than enforcing s(C)PC. As our first experimental attempts (not presented in this paper) to enforce s(C)2SAC show that this is a very expensive approach, we do believe that s(C)DC is, for now, the best second-order consistency to be enforced at preprocessing.

A second unsurprising lesson is that there are problems for which enforcing s(C)DC is not cost-effective. Basically, the cost of enforcing s(C)DC mainly depends on the total number of values to be tested[9] as well as on the time complexity of the underlying GAC algorithm(s). For non-binary constraints, the time complexity of enforcing GAC usually increases with the arity of the constraints and for extensional constraints, the time complexity of enforcing GAC usually depends on the size of the tables. As a consequence, when a problem involves constraints of large arity and/or large tables, enforcing s(C)DC may become penalizing. This is the case for the `renault-mod`, `tsp-20` and `tsp-25` series in Table 3.

A third lesson is that enforcing sDC at preprocessing tends to make the MAC algorithm more robust. To confirm this, Figure 16 shows two cactus-shaped plots that give some insight on the relative performance of the variants of MAC on the whole range of tested problem instances. These plots show that establishing sDC before searching for a solution using MAC enhances the robustness of the solver, especially when the variable ordering heuristic fails. With the *dom/ddeg* variable ordering heuristic, plain MAC is almost always worse than MAC with some second-order consistency established during the preprocessing. With the better *dom/wdeg* variable ordering heuristic, enforcing a second-order consistency is interesting for the hardest problems, those that require more than 50 seconds to be solved. A zoom in Figure 16(b) on MAC versus sDC1-MAC shows how the trend is reversed: after 50 seconds, sDC1-MAC solves more instances than MAC.

We conclude this section with an experiment on the hardest instances from the RLFAP series `scens11` (see e.g., the results at `http://www.cril.fr/{CPAI06,CPAI08,CPAI09}`). Using the variable ordering heuristic *dom/wdeg*, we ran plain MAC, MAC with the symmetry-breaking method described by Lecoutre & Tabary (2009, denoted MAC$_{SB}$ here), s(C)DC-MAC, and finally MAC with both inference mechanisms. Enforcing sCDC permits to reduce the size of the search tree developed by MAC on these instances, but for sDC, this is less obvious (the added constraints perturbate the heuristic), as shown by Table 5. Interestingly, the joint use of the symmetry-breaking method reveals to be quite efficient here. Using both inference mechanisms, the unsatisfiability of all instances can be proved without any search effort (0 nodes in the sCDC1-MAC$_{SB}$ and sDC1-MAC$_{SB}$ columns). Note that SAC-MAC$_{SB}$ builds a search tree for these problems, and it is not even possible to solve the instance `scen11-f1` with SAC1-MAC$_{SB}$ (or SAC3-MAC$_{SB}$) equipped with the heuristic *dom/ddeg*, within 20 minutes.

---

9. For example, we discarded the `fapp` series of problem instances, for which consistencies based on singleton checks are clearly not adapted because of the huge number of values.

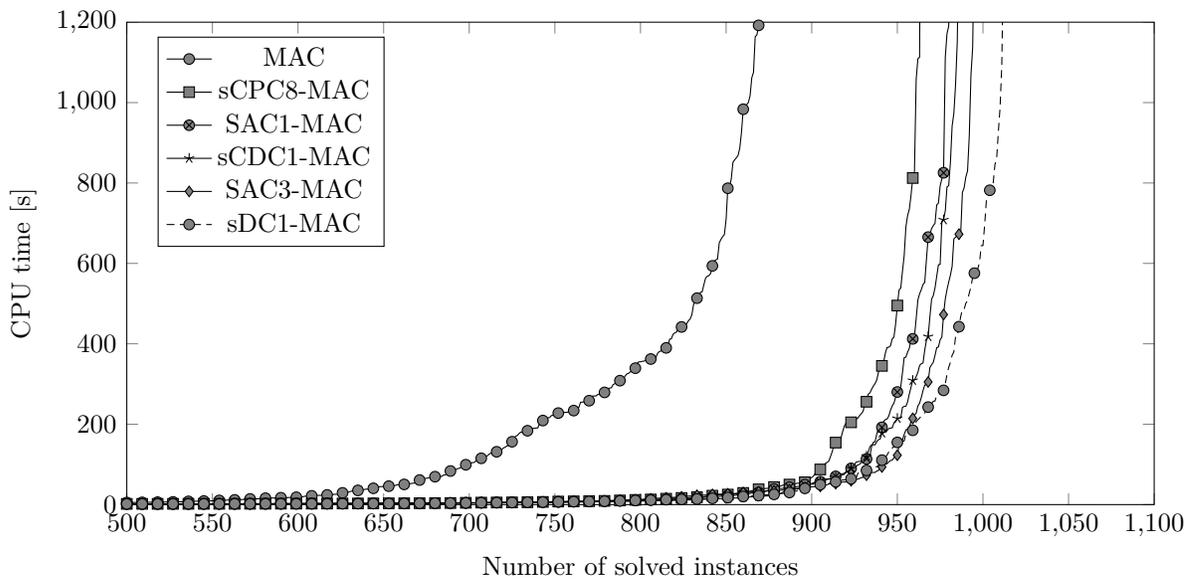

(a) with *dom/ddeg*

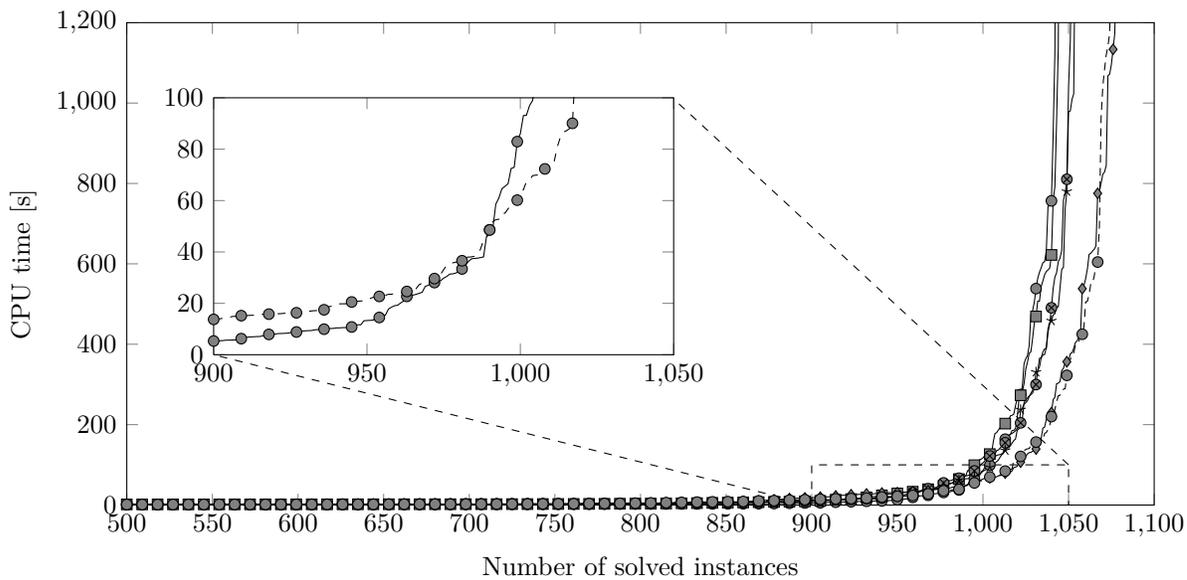

(b) with *dom/wdeg*

Figure 16: Number of instances (out of the full set of 1,237 instances) that can be solved within a given amount of CPU time with variants of MAC (e.g., 870 instances can be solved within 1,200 s by MAC-*dom/ddeg*).

| Instance | | MAC | MAC$_{SB}$ | sDCDC1- | | sDC1- | |
|---|---|---|---|---|---|---|---|
| | | | | MAC | MAC$_{SB}$ | MAC | MAC$_{SB}$ |
| `scen11-f1` | cpu | > 1,200 | 52.8 | > 1,200 | 10.0 | > 1,200 | 12.3 |
| | nodes | | 267 K | | 0 | | 0 |
| `scen11-f2` | cpu | > 1,200 | 24.3 | > 1,200 | 7.6 | > 1,200 | 9.7 |
| | nodes | | 108 K | | 0 | | 0 |
| `scen11-f3` | cpu | > 1,200 | 11.0 | > 1,200 | 5.8 | > 1,200 | 6.8 |
| | nodes | | 35,979 | | 0 | | 0 |
| `scen11-f4` | cpu | 542 | 7.9 | 438 | 6.0 | 808 | 5.9 |
| | nodes | 3,381 K | 11,246 | 2,095 K | 0 | 3,362 K | 0 |
| `scen11-f6` | cpu | 60.1 | 3.8 | 41.9 | 4.3 | 65.1 | 4.6 |
| | nodes | 348 K | 2,226 | 163 K | 0 | 215 K | 0 |
| `scen11-f8` | cpu | 6.0 | 3.8 | 9.94 | 5.3 | 11.1 | 4.3 |
| | nodes | 14,077 | 1,847 | 5,021 | 0 | 4,518 | 0 |

Table 5: Cost of running (Φ-)MAC-*dom/wdeg* on the hardest instances of the RLFAP series `scens11`. MAC$_{SB}$ is MAC with an automatic global symmetry-breaking method.

## 7. Conclusion

This paper is intended to give a better picture of second-order consistencies. For this purpose, we have studied the theoretical relationships existing between four basic second-order consistencies (and their variants), and we have shown that some of them can be reasonably enforced *before* search. However, in the next generation of constraint solvers, tractable classes of CSP instances will certainly have to be identified and exploited *during* search, in order to close, for example, certain nodes of the search tree in polynomial time. Because several theoretical results relate global consistency to second-order consistencies (e.g., strong 3-consistency), this should increase the importance of second-order consistencies.

In practical terms, there are some advantages of using (conservative) dual consistency. Algorithms to enforce strong (C)DC are rather easy to implement, and made efficient because of highly optimized underlying GAC algorithms. Used at preprocessing, they reveal to improve the robustness of a constraint solver on several hard structured problems. (C)2SAC is stronger than (C)DC, but a naive approach for establishing it requires several passes of $O(n^2d^2)$ enforcements of GAC, which makes it ineffective. A perspective of this work is to devise efficient algorithms for (C)2SAC that could be competitive with (C)DC ones.

Finally, as multi-core processors become increasingly common, parallel constraint solving should become more and more useful. In the near future, we may imagine that strong second-order consistencies could be used as basic components (with strong inference capabilities) of parallel solvers.

## Acknowledgments

This paper is an extended revised version of earlier works (Lecoutre et al., 2007a, 2007b). We would like to thank the anonymous reviewers for their constructive remarks.